\documentclass[5p,twocolumn]{elsarticle}
\makeatletter
\def\ps@pprintTitle{%
	\let\@oddhead\@empty
	\let\@evenhead\@empty
	\def\@oddfoot{\rightline{\thepage}}%
	\let\@evenfoot\@oddfoot}
\makeatother

\usepackage{lineno,hyperref}
\usepackage[justification=centering]{caption} 
\usepackage[export]{adjustbox}
\usepackage{subcaption}
\usepackage{amsmath}
\usepackage{amsfonts}
\usepackage[boxed]{algorithm2e}
\usepackage{booktabs}
\usepackage{multirow}
\usepackage{color,soul}  
\usepackage{mathtools, nccmath}
\usepackage{amssymb}
\usepackage{pifont}
\usepackage{changepage}

\newcommand{\cmark}{\ding{51}}
\newcommand{\xmark}{\ding{55}}

\DeclarePairedDelimiter{\nint}\lfloor\rceil

\soulregister\cite7      
\soulregister\ref7		 
\modulolinenumbers[5]

\begin{document}

\begin{frontmatter}

\title{Motion-Encoded Particle Swarm Optimization \\
	for Moving Target Search Using UAVs}

\author[1,2]{Manh Duong Phung\corref{cor1}}
\ead{manhduong.phung@uts.edu.au}

\author[1]{Quang Phuc Ha}
\ead{quang.ha@uts.edu.au}

\address[1]{School of Electrical
	and Data Engineering, University of Technology Sydney (UTS) \\
	15 Broadway, Ultimo NSW 2007, Australia}
\address[2]{VNU University of Engineering and Technology (VNU-UET), Vietnam National University, Hanoi (VNU) \\ 144 Xuan Thuy, Cau Giay, Hanoi, Vietnam}

\cortext[cor1]{Corresponding author}

\begin{abstract}
This paper presents a novel algorithm named the motion-encoded particle swarm optimization (MPSO) for finding a moving target with unmanned aerial vehicles (UAVs). From the Bayesian theory, the search problem can be converted to the optimization of a cost function that  represents the probability of detecting the target. Here, the proposed MPSO is developed to solve that problem by encoding the search trajectory as a series of UAV motion paths evolving over the generation of particles in a PSO algorithm. This motion-encoded approach allows for preserving important properties of the swarm including the cognitive and social coherence, and thus resulting in better solutions. Results from extensive simulations with existing methods show that the proposed MPSO improves the detection performance by 24\% and time performance by 4.71 times compared to the original PSO, and moreover, also outperforms other state-of-the-art metaheuristic optimization algorithms including the artificial bee colony (ABC), ant colony optimization (ACO), genetic algorithm (GA), differential evolution (DE), and tree-seed algorithm (TSA) in most search scenarios. Experiments have been conducted with real UAVs in searching for a dynamic target in different scenarios to demonstrate MPSO merits in a practical application.\\
\end{abstract}

\begin{keyword}
Optimal search \sep Particle swarm optimization \sep UAV \\
\textit{\textbf{Source code:}} The implementation of MPSO can be found at \href{https://github.com/duongpm/MPSO}{\textbf{https://github.com/duongpm/MPSO}}
\end{keyword}

\end{frontmatter}


\section{Introduction}
Unmanned aerial vehicles (UAVs) have been receiving much research interest with numerous practical applications, especially in surveillance and rescue due to their capability of operating in harsh environments with sensor-rich work capacity suitable for different tasks. In searching for a lost target using UAVs, there often exists a critical period called ``golden time" in which the probability the target being found should be highest \cite{Sergio2015}. As time progresses, that probability rapidly decreases due to the attenuation of initial information and the influence of external factors such as weather conditions, terrain features and target dynamics. The main objective in searching for a lost target using UAVs therefore includes finding a path that can maximize the probability of detecting the target within a specific flight time given initial information on target position and search conditions \cite{Bourgault2006, Raap2017}.

In the literature, the search problem is often formulated as probabilistic functions so that uncertainties in initial assumptions, search conditions and sensor models can be adequately incorporated. In \cite{Bourgault2006, Lanillos2013}, a Bayesian approach has been introduced to derive the objective functions for evaluating the detection probability of UAV flight paths. The initial search map has been modeled as a multivariate normal distribution with the mean and variance being computed based on initial information about the target position \cite{Sara2018, Furukawa2006}. In \cite{Raap2017, Furukawa2006}, the target dynamic is represented by a stochastic Markov process which can then be deterministic or not depending on the searching scenarios. The sensor, on the other hand, is often modeled as either a binary variable with two states, ``detected" or ``not detected" \cite{Sara2018}, or as a continuous Gaussian variable \cite{Bourgault2006}. 

Due to various probabilistic variables involved, the complexity of the searching problem varies from the level of nondeterministic polynomial-time hardness (NP-hard \cite{Trummel1986}) to nondeterministic exponential-time completeness (NEXP-complete \cite{Bernstein2002}), in which the number of solutions available to search grows exponentially with respect to the search dimension and flight time. Consequently, solving this problem using classical methods such as differential calculus to find the exact solution becomes impractical, and hence, approximated methods are often used. A number of methods have been developed, such as greedy search with one-step look ahead \cite{Bourgault2006} and k-step look ahead \cite{Raap2017}, ant colony optimization (ACO) \cite{Sara2018}, Bayesian optimization approach (BOA) \cite{Lanillos2013}, genetic algorithm (GA) \cite{Goldberg1989,Lin2009}, cross entropy optimization (CEO) \cite{Lanillos2012}, branch and bound approach \cite{Eagle1990}, limited depth search \cite{Alejandro2009}, and gradient descend methods \cite{Gan2011,Mathews2007}. Table \ref{tab:review} compares main properties of some algorithms where the ``multi-agent" column implies the possibility of using multiple UAVs for searching and ``ad hoc heuristic" for the case being specifically designated for the search problem. It is noted that most methods cope with moving targets and use the binary model for detection sensors. Some approaches (\cite{Lanillos2013,Sara2018,Lanillos2012,Alejandro2009}) employ multiple UAVs to speed up the search process, whereas others use ad hoc heuristic to improve detection probability. 

\begin{table*}[]
	\centering
	\caption{Comparison between search methods}
	\label{tab:review}
	\begin{tabular}{cccccc}
		\hline
		\rule{0pt}{3ex}
		{Method}                                                 & {Work} & {Target} & {\begin{tabular}[c]{@{}c@{}}Binary\\   sensor\end{tabular}} & {\begin{tabular}[c]{@{}c@{}}Multi-\\   agent\end{tabular}} & {\begin{tabular}[c]{@{}c@{}}Ad hoc\\   heuristic\end{tabular}} \\
		\hline
		\begin{tabular}[c]{@{}c@{}}\small{one-step look ahead}\end{tabular} & \cite{Bourgault2006} &              
		\begin{tabular}{@{}c@{}}\small{Static \& Dynamic}  \end{tabular}
		& \xmark                                                                & \xmark                                                                & \cmark                                                       \\
		\begin{tabular}[c]{@{}c@{}}\small{k-step look ahead}\end{tabular}   & \cite{Raap2017}             & Dynamic         & \cmark                                                                  & \xmark                                                                & \cmark                                                                     \\
		BOA                                                             & \cite{Lanillos2013}             & Dynamic         & \cmark                                                                  & \cmark                                                                & \xmark                                                                     \\
		ACO                                                             & \cite{Sara2018}             & Dynamic         & \cmark                                                                  & \cmark                                                                & \cmark                                                                     \\
		GA                                                              & \cite{Lin2009}            & Static          & \cmark                                                                  & \xmark                                                                & \xmark                                                                     \\
		CEO                                                             & \cite{Lanillos2012}            & Dynamic         & \cmark                                                                  & \cmark                                                                & \xmark                                                                     \\
		Depth search                                                    & \cite{Alejandro2009}            & Static          & \cmark                                                                  & \cmark                                                                & \cmark                                                            \\
		Gradient descent                                                & \cite{Gan2011}            & Static          & \xmark                                                                  & \xmark                                                                & \xmark                                     \\                              
		\bottomrule		 
	\end{tabular}
\end{table*}

From the literature, it is recognizable that approaches to optimal search diverge in assumptions, constraints, target dynamics and searching mechanisms. Due to its complex nature, optimal search, especially in scenarios with fast-moving targets, remains a challenging problem. Besides, recent advancements in sensor, communication and UAV technologies enable the development of new search platforms. They pose the need for new methods that should not only robust in search capacity but also possess properties such as computational efficiency, adaptability and optimality.

For optimization, particle swarm optimization (PSO) is a potential technique with a number of key advantages that have been successfully applied in various applications \cite{Kennedy2001,Lee2006,MOHAMMADI2018,Niknam2013,Niknam2012}. It is less sensitive to initial conditions as well as the variation of objective functions and is able to adapt to many search scenarios via a small number of parameters including an initial weight factor and two acceleration coefficients \cite{Eberhart1998}. It generally can find the global solution with a stable convergence rate and shorter computation time compared to other stochastic methods \cite{Gaing2003}. More importantly, PSO is simple in implementation with the capability of being parallelized to run with not only computer clusters or multiple processors but also graphical processing units (GPU) of a single graphical card. This allows to significantly reduce the execution time without requiring any change to the system hardware \cite{PHUNG2017}.

Motivated from the aforementioned analysis, we will employ the PSO methodology in this study to deal with the search problem in complex scenarios for fast moving targets, aiming to improve the search performance in both detection probability and execution time. To this end, we propose a new motion-encoded PSO algorithm, taking into account both cognitive and social coherence of the swarm. Our contributions include: (i) the formulation of an objective function for optimization, incorporating all assumptions and constraints, from the search problem and the probabilistic framework; (ii) the development of a new motion-encoded PSO (MPSO) from the idea of changing the search space for the swarm to avoid getting stuck at local maxima; (iii) the demonstration of MPSO implemented for UAVs in experimental search scenarios to validate its outperformance over other PSO algorithms obtained from extensive comparison analysis. The results show that MPSO, on one hand, presents superior performance on various search scenarios while on the other hand remains simple for practical implementation. 

The rest of this paper is structured as follows. Section~\ref{sect:formulation} outlines the steps to formulate the objective function. Section~\ref{sect:PSO} presents the proposed MPSO and its implementation for solving a complex search problem. Section~\ref{sect:experiment} provides simulation and experimental results. A conclusion is drawn in Section~\ref{conclusion} to close our paper.

\section{Problem Formulation}
\label{sect:formulation}

The search problem is formulated by modeling the target, sensor and belief map with details as follows.

\subsection{Target Model}
In the searching problem, the target is described by an unknown variable $x \in X$ representing its location. Before the search starts, a probability distribution function (PDF) is used to model the target location based on the available information, e.g., the last known location of the target before losing its signal. This PDF could be a normal distribution centered about the last known location, but also could be a uniform PDF if nothing is known about the target location. In the searching space, this PDF is represented by a grid map called the belief map, $b(x_0)$, in which the value in each cell corresponds to the probability of the target being in that cell. The map can be created by discretizing the searching space $S$ into a grid of $S_r \times S_c$ cells and associating a probability to each cell. Assume the target presents in the searching space, we have $\sum_{x_0 \in S} b(x_0)= 1$.

During the searching process, the target may be not static but navigate in a certain pattern. This pattern can be modeled by a stochastic process which can be assumed as a Markov process. In the special case of a conditionally deterministic target, which is considered in this study, that pattern merely depends on the initial position $x_0$ of the target. In that case, the transition function, $p(x_t|x_{t-1})$, representing the probability which the target goes from cell $x_{t-1}$ to $x_t$, is known for all cells $x_t \in S$. Consequently, the path of the target will be entirely known if its initial position is known. This assumption is made quite often for the survivor search at sea \cite{iida1998} and also for the search problems in general \cite{Sara2018}.

\subsection{Sensor Model}
In order to look for and find a target, a sensor is installed on the UAV to carry out an observation $z_t$ at each time step $t$. The observations are independent such that the occurrence of one observation provides no information about the occurrence of the other observation. A detection algorithm is implemented to return a result for each observation which is assumed to have only two possible outputs, the detection of the target,  $z_t = D_t$, or no detection, $z_t = \bar{D_t}$, where $D_t$ represents a ``detection" event at time $t$. Due to imperfectness of the sensor and detection algorithm, an observation of the target detected, $z_t = D_t$, still does not ensure the presence of the target at $x_t$. This is reflected through the observation likelihood, $p(z_t|x_t)$, given knowledge of the sensor model. The likelihood of no detection, given a target location $x_t$, is then computed by:

\begin{equation}\label{eq:sensor}
p(\bar{D_t}|x_t) = 1 - p(D_t|x_t).
\end{equation} 

\subsection{Belief Map Update}
Once the initial distribution, $b(x_0)$, is initialized, the belief map of the target at time $t$, $b(x_t)$, can be established based on the Bayesian approach and the sequence of observations, $z_{1:t}=\{z_1,...,z_t\}$, made by the sensor. This approach is conducted recursively via two phases, prediction and update. In the prediction, the belief map is propagated over time in accordance with the target motion model. Suppose at time $t$, the previous belief map, $b(x_{t-1})$, is available. Then, the predicted belief map is calculated as:

\begin{equation}\label{eq:predict}
\hat{b}(x_t) = \sum_{x_{t-1} \in S} p(x_t|x_{t-1}) b(x_{t-1}).
\end{equation} 

Notice from (\ref{eq:predict}) that the belief map $b(x_{t-1})$ is in fact the conditional probability of the target being at $x_{t-1}$ given observations up to $t-1$, $b(x_{t-1}) = p(x_{t-1}|z_{1:t-1})$. When the observation $z_t$ is available, the update is conducted simply by multiplying the predicted belief map by the new conditional observation likelihood as follows:

\begin{equation}\label{eq:update}
b(x_t) = \eta_t p(z_t|x_t) \hat{b}(x_t) ,
\end{equation}    
where $\eta_t$ is the normalization factor,
\begin{equation}\label{eq:normfactor}
\eta_t = 1/\sum_{x_t \in S} p(z_t|x_t) \hat{b}(x_t).
\end{equation} 
$\eta_t$ scales the probability that the target presents inside the searching area to one, i.e., $\sum_{x_t \in S} b(x_t) = 1$.

\subsection{Searching Objective Function}
According to the Bayesian theory, the probability that the target does not get detected at time $t$ during an observation, $r_t = p(\bar{D}_t|z_{1:t-1})$, relies on two factors: (i) the latest belief map from the prediction phase (\ref{eq:predict}), and (ii) the no detection likelihood (\ref{eq:sensor}). Across the whole searching area, that probability is given by:

\begin{equation}\label{eq:nodetection}
r_t = \sum_{x_t \in S} p(\bar{D}_t|x_t) \hat{b}(x_t).
\end{equation} 
Notice that $r_t$ is exactly the inverse of the normalization factor $\eta_t$ in (\ref{eq:normfactor}), $r_t = 1/\eta_t$, for a ``no detection" event, $z_t = \bar{D_t}$, and thus is smaller than 1. By multiplying the not detected probability $r_t$ over time, the joint probability of failing to detect the target from time 1 to $t$, $R_t =p(\bar{D}_{1:t})$,  is then obtained:

\begin{equation}\label{eq:allNoDetection}
R_t = \prod_{k=1}^{t} r_{k} = R_{t-1}r_t.
\end{equation} 
Hence, the probability that the target gets detected for the first time at time $t$ is computed as:

\begin{equation}\label{eq:firstDetection}
p_t = \prod_{k=1}^{t-1} r_k(1-r_t) = R_{t-1}(1-r_t).
\end{equation} 
Summing $p_t$ over $t$ steps gives the probability of detecting the target in $t$ steps:

\begin{equation}\label{eq:AllDetection2}
P_t = \sum_{k=1}^{t} p_k = P_{t-1} + p_t.
\end{equation}
$P_t$ is thus often referred to as the ``cumulative" probability to distinguish it with $p_t$. Notice that

\begin{equation}\label{eq:allDetection}
P_t = 1 - R_t,
\end{equation} 
and as $t$ grows, the probability of first detection $p_t$ becomes smaller because the chance of detecting the target in previous steps increases. The cumulative probability $P_t$ is thus bounded and increases toward one as $t$ goes to infinity.

The objective function for the searching problem can now be formulated based on (\ref{eq:AllDetection2}) given a finite search time. Let the search time period be $\{1, . . . , N\}$, the goal of the searching strategy is to determine a search path $O = (o_1, . . . , o_N )$ that could maximize the cumulative probability $P_t$. As such, the objective function is eventually formulated as follows:

\begin{equation}\label{eq:J1}
J = \sum_{t=1}^{N} p_t.
\end{equation} 

\section{Motion-encoded Particle Swarm Optimization}
\label{sect:PSO}
As the search problem defined in (\ref{eq:J1}) is NP-hard \cite{Trummel1986,Bernstein2002}, the time required to calculate all possible paths to find the optimal solution would greatly increase and become intractable. Therefore, a heuristic approach like PSO can be a good option for solving the optimal search problem as in this study.

\subsection{Particle Swarm Optimization}
PSO is a population-based stochastic technique, inspired by social behavior of bird flocking, designed for solving optimization problems \cite{Kennedy2001, Kohler2019}. In PSO, a swarm of particles is initially generated with random positions and velocities. Each particle then moves and evolves in a cognitive fashion with other particles to seek the global optimum. Those movements are driven by its best position, $L_k$, and the best position of the swarm, $G_k$. Let $x_k$ and $v_k$ be the position and velocity of a particle at generation $k$, respectively. The movement of that particle in the next generation is given by:
\begin{equation}
\label{eq:velocity}
v_{k+1} \leftarrow wv_k + \varphi_1r_1(L_k - x_k) + \varphi_2r_2(G_k - x_k)
\end{equation}
\begin{equation}
\label{eq:position}
x_{k+1} \leftarrow x_k + v_{k+1} ,
\end{equation}
where $w$ is the inertial weight, $\varphi_1$ is the cognitive coefficient, $\varphi_2$ is the social coefficient, and $r_1$, $r_2$ are random sequences sampled from a uniform probability distribution in the range [0,1]. From (\ref{eq:velocity}) and (\ref{eq:position}), the movement of a particle is directed by three factors, namely, following its own way, moving toward its best position, or moving toward the swarm's best position. The ratio among those factors is determined by the values of $w$, $\varphi_1$, and $\varphi_2$. 

\subsection{MPSO for Optimal Search}
There have been several modifications and improvements from the PSO algorithm, depending on the application. However, the implementation of PSO for online searching for dynamic targets in a complex environment remains a challenging task, particularly in a limited time window. For the search problem, it is desired to encode the position of particles in a way that the particles can gradually move toward the global optimum. A common approach is to define a position as a multi-dimensional vector representing a possible search path:
\begin{equation} \label{eq:particle}
x_k \sim O_k = (o_{k,1}, ... , o_{k,N}),
\end{equation}
where $o_{k,i}$ corresponds to a node of the search map \cite{Roberge2013,ZHANG2013}. The drawback of this approach is that it does not cover the adjacent dynamic behavior in path nodes and thus may result in invalid paths during the searching process. Discrete PSO can be used to overcome this problem, but the momentum of particles is not preserved, causing local maxima \cite{Clerc2004}. Indirect approaches such as the angle-encoded PSO \cite{Fu2012} and priority-based
encoding PSO \cite{Mohemmed2008} can be a good option to deal with it and generate better results. Their mapping functions, however, require the phase angles to be within the range of $[-\pi/2,\pi/2]$ which limits the search capacity, especially in a large dimension.

\begin{figure}
	\centering
	\includegraphics[width=0.9\linewidth]{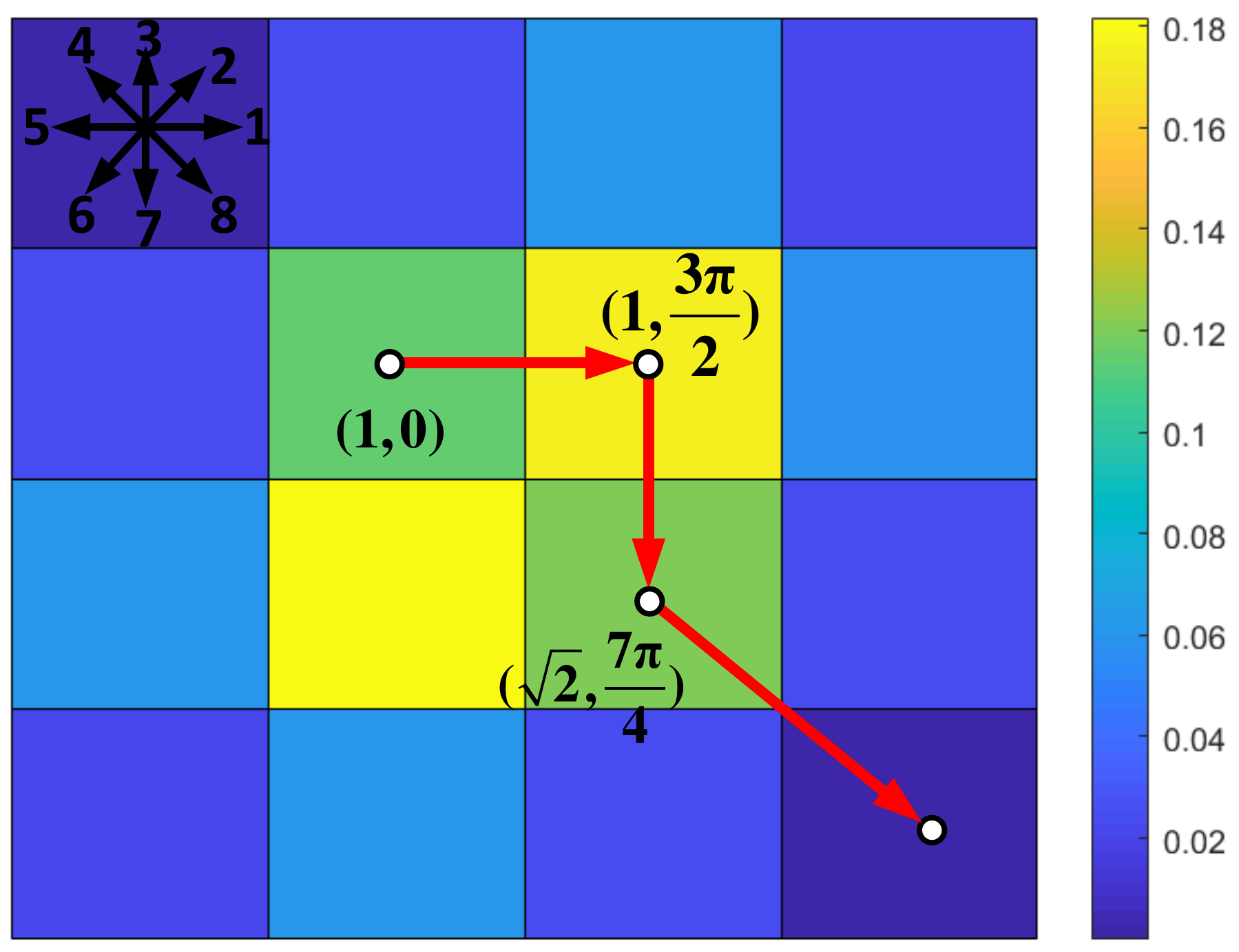}
	\caption{Motion-encoded illustration for a path with three segments, $U_k = ((1,0),(1,3\pi/2), (\sqrt{2},7\pi/4))$}
	\label{fig:motionEncode}
\end{figure}

Here, we propose the idea of using UAV motion to encode the position of particles. Instead of using nodes, we view each search path as a set of UAV motional segments, each corresponds to the movement of UAV from its current cell to another on the plane of flight. By respectively defining the magnitude and direction of the motion at time $t$ as $\rho_t$ and  $\alpha_t$, that motion can be completely described by a vector $u_t=(\rho_t,\alpha_t)$.  A search path is then described by a vector of $N$ motion segments, $U_k = (u_{k,1}, ... , u_{k,N})$. Using $U_k$ as the position of each particle, equations for MPSO can be written as:

\begin{equation}
\label{eq:velocity2}
\Delta U_{k+1} \leftarrow wU_k + \varphi_1r_1(L_k - U_k) + \varphi_2r_2(G_k - U_k)
\end{equation}

\begin{equation}
\label{eq:position2}
U_{k+1} \leftarrow U_k + \Delta U_{k+1}.
\end{equation}
Figure \ref{fig:motionEncode} illustrates a path with three segments, $U_k = ((1,0),(1,3\pi/2), (\sqrt{2},7\pi/4))$,  where the belief map is colour-coded with probability values indicated on the right. 

During the search, it is also required to map $U_k$ to a direct path $O_k$ so that the cost associated with $U_k$ can be evaluated. As shown in Fig. \ref{fig:motionEncode}, the mapping process can be carried out by first constraining the UAV motion to one of its eight neighbors in each time step. Then, the motion magnitude $\rho_t$ can be normalized and the motion angle $\alpha_t$ can be quantized as:

\begin{equation}
\label{eq:magnitude}
\rho_t^* = 1 
\end{equation}

\begin{equation}
\label{eq:rounding}
\alpha_t^* = 45 ^ {\circ} \nint{\alpha_t / 45 ^ {\circ}}, 
\end{equation}
where $\nint{}$ represents the operator for rounding to the nearest integer. Node $o_{k,t+1}$ corresponding to the location of UAV in the Cartesian space is then given by:

\begin{equation}
\label{eq:mapping}
o_{k,t+1} = o_{k,t} + u_{k,t}^* ,
\end{equation} 
where
\begin{equation}
\label{eq:coordinate}
u_{k,t}^* = (\nint{cos{\alpha_t^*}},\nint{sin{\alpha_t^*}}).
\end{equation} 

From the decoded path $O_k$, the cost value can be evaluated by the objective function (\ref{eq:J1}) and then the local and global best can be computed as follows:

\begin{equation} \label{eq:LBest}
L_k = \left\{\begin{array}{ll}
U_k \quad \text{if $J(O_k) > J(L^*_{k-1})$}\\
L_{k-1} \quad \mathrm{otherwise}
\end{array}\right., 
\end{equation}

\begin{equation} \label{eq:GBest}
G_k = \underset{L_k}{\operatorname{argmax}}J(O_k),
\end{equation}
where $L^*_{k}$ is the decoded path of $L_{k}$. It can be seen from the mapping process that (\ref{eq:rounding}) discretizes the motion to one of eight possible directions, (\ref{eq:coordinate}) converts the moving direction to an increment in Cartesian coordinates, and (\ref{eq:mapping}) incorporates the increment to form the next node of the path. 

Similarly to the interchange between the time domain and frequency domain in signal analysis, the mapping process of MPSO allows particles to search in the motion space instead of the Cartesian space. This leads to the following advantages:

\begin{itemize}
\item  The motion space maintains the location of nodes consecutively so that the resultant paths after each generation evolvement are always valid, which is not the case of the Cartesian space;
\item In motion space, the momentum of particles and swarm behaviors including exploration and exploitation are preserved so that the search performance is maintained and the swarm is able to cope with different target dynamics;
\item As the normalization of $\rho_t$ and quantization of $\alpha_t$ in (\ref{eq:magnitude}) and (\ref{eq:rounding}) are only carried out for the purpose of cost evaluation, their continuous values are still being used for velocity and position updates as in (\ref{eq:velocity2}) - (\ref{eq:position2}). This property is important to avoid the discretizing effect of PSO so that the search resolution is not affected.
\end{itemize}

Finally, it is also noted that MPSO preserves the search mechanism of PSO via its update equations (\ref{eq:velocity2}) - (\ref{eq:position2}) so that the advantages of PSO such as stable convergence, independence of initial conditions and implementation feasibility can be maintained. 

\subsection{Implementation}
\begin{figure}
	\centering
	\includegraphics[width=0.6\linewidth]{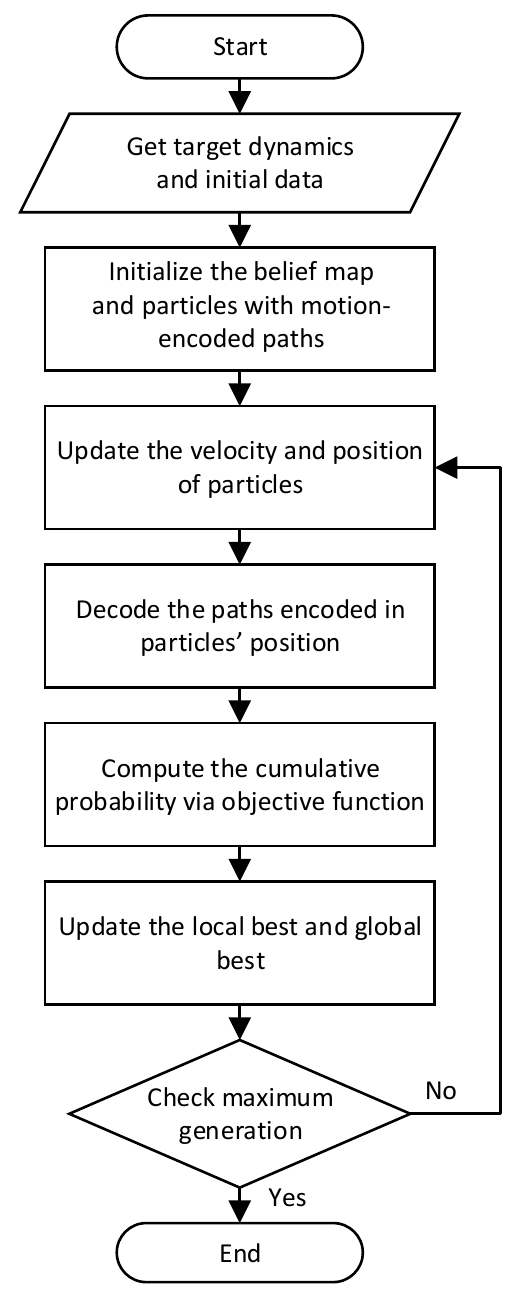}
	\caption{Flowchart of MPSO algorithm}
	\label{fig:flowchart}
\end{figure}

Figure \ref{fig:flowchart} shows the flowchart of MPSO to illustrate the implementation presented in Algorithm \ref{fig:pseudocode}. Its structure is based on the core PSO but extended with the incorporation of the motion encoding and decoding steps. The belief map update as in (\ref{eq:predict}) and (\ref{eq:update}) needs to be conducted during calculating the fitness when the target is non-static. Notably, the parallelism technique proposed in \cite{PHUNG2017} can be applied to speed up the computation process of MPSO.

\section{Results}
\label{sect:experiment}
To evaluate the performance of MPSO, we have conducted extensive simulation, comparison and experiments with detail described below.

\begin{algorithm}
	\caption{Pseudo code of MPSO.}
	\label{fig:pseudocode}
	\tcc{Initialization:}
	Get target dynamics and initial data\;
	Create belief map\;
	Set swarm parameters $w$, $\varphi_1$, $\varphi_2$, $swarm\_size$\; 
	\ForEach {particle in swarm} {
		\quad	Create random motion-encoded paths $U_k$\;
		\quad	Assign $U_k$ to particle position\;
		\quad	Compute $fitness$ value of each particle\;
		\quad	Set $local\_best$ value of each particle to itself\; 
		\quad	Set $velocity$ of each particle to zero\; 
	}
	Set $global\_best$ to the best fit particle\;
	\tcc{Evolutions:}
	\For{$k \gets 1$ to $max\_generation$} {
		\ForEach {particle in swarm} { 
			Compute motion velocity $\Delta U_{k+1}$; \tcc*[f]{Eq.\ref{eq:velocity2}} \\
			Compute new position $U_{k+1}$; \tcc*[f]{Eq.\ref{eq:position2}} \\
			Decode $U_{k+1}$ to $O_{k+1}$; \tcc*[f]{Eq.\ref{eq:coordinate} - \ref{eq:mapping}} \\
			Update $fitness$ of $O_{k+1}$; \tcc*[f]{Eq.\ref{eq:J1}} \\
			Update $local\_best$ $L_{k+1}$; \tcc*[f]{Eq.\ref{eq:LBest}} \\
		}
		Update $global\_best$ $G_{k+1}$; \tcc*[f]{Eq.\ref{eq:GBest}} \\
	}
\end{algorithm}

\subsection{Scenarios setup}
\begin{figure*}
	
	\begin{subfigure}{0.5\textwidth}
		\centering
		\includegraphics[width = \textwidth]{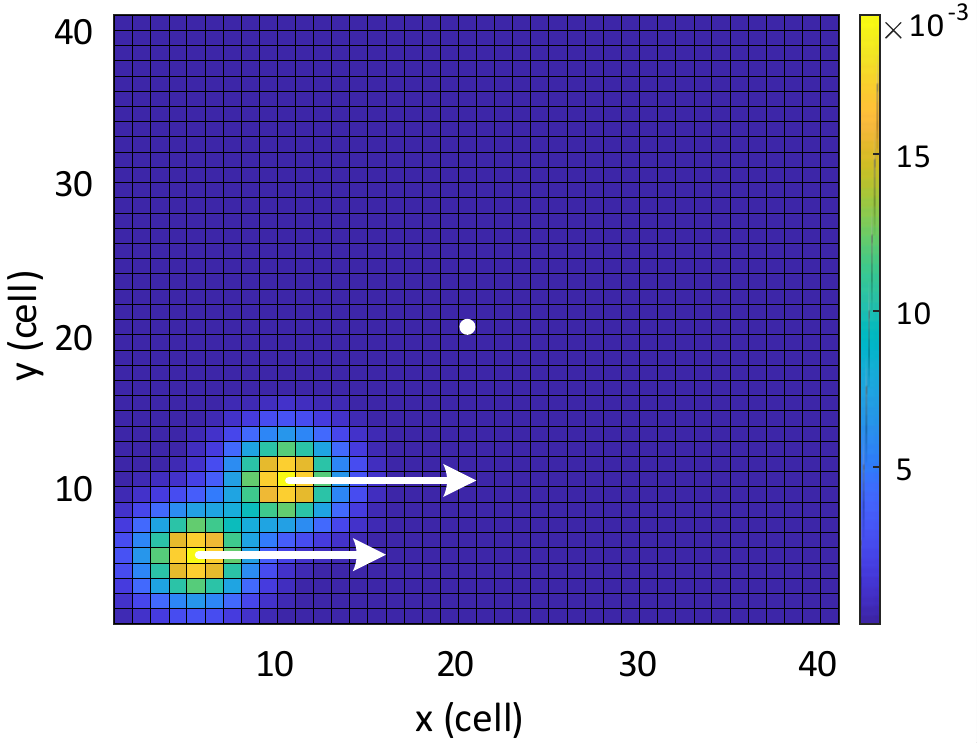}
		\caption{Scenario 1}
		\label{fig:scenario1}
	\end{subfigure}%
	\begin{subfigure}{0.5\textwidth}
		\centering
		\includegraphics[width = \textwidth]{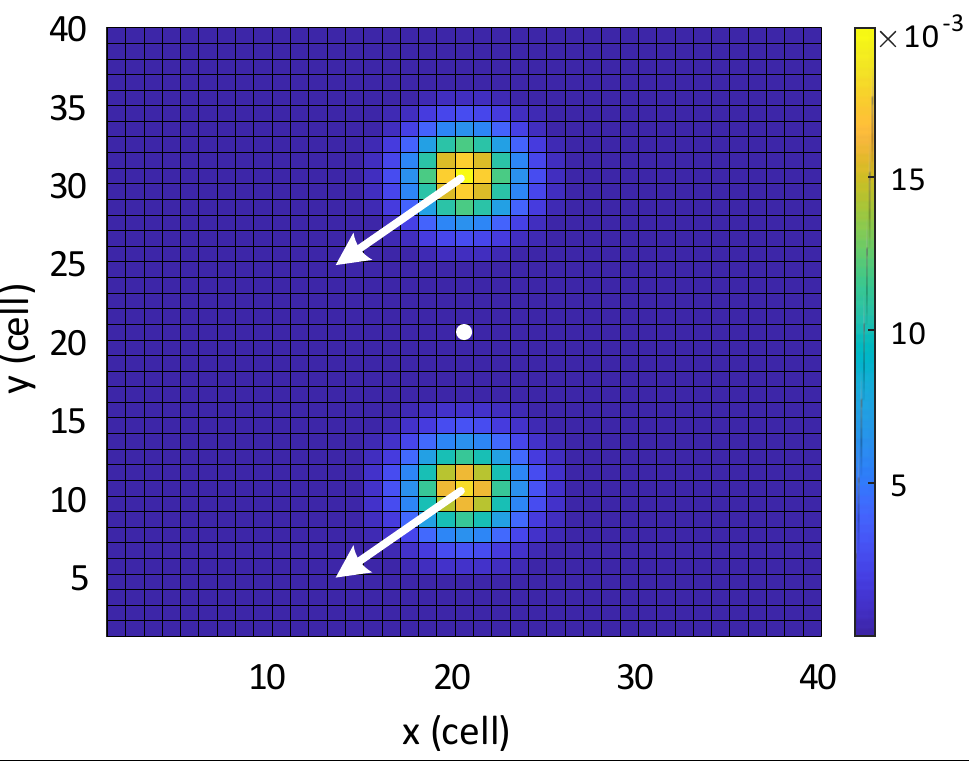}
		\caption{Scenario 2}
		\label{fig:scenario2}
	\end{subfigure}
	\begin{subfigure}{0.5\textwidth}
		\centering
		\includegraphics[width = \textwidth]{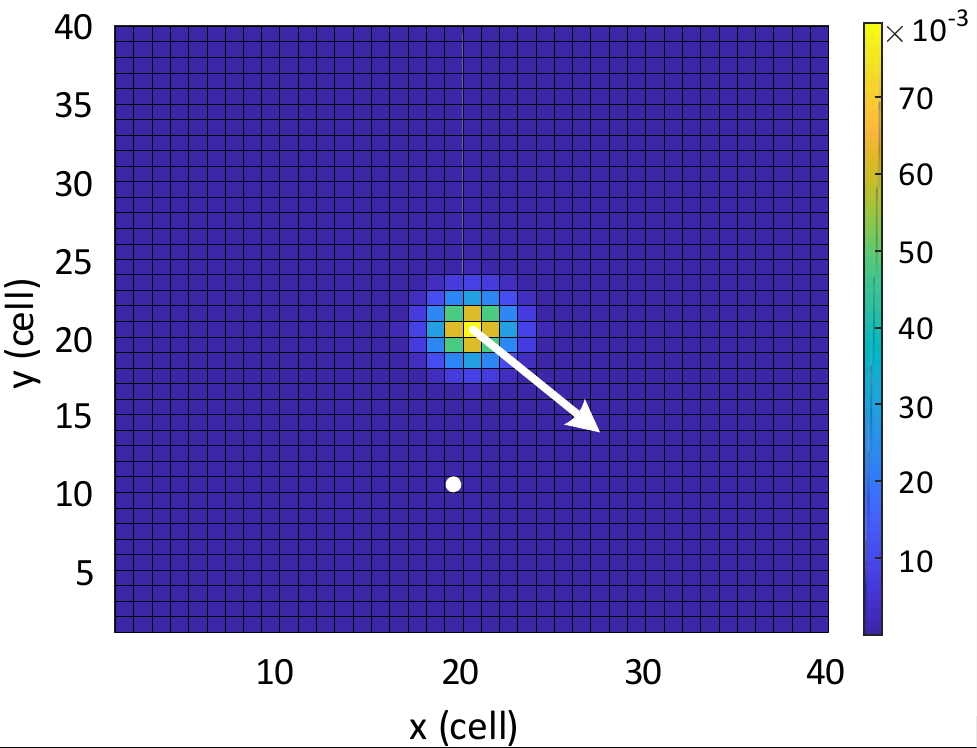}
		\caption{Scenario 3}
		\label{fig:scenario3}
	\end{subfigure}%
	\begin{subfigure}{0.5\textwidth}
		\centering
		\includegraphics[width = \textwidth]{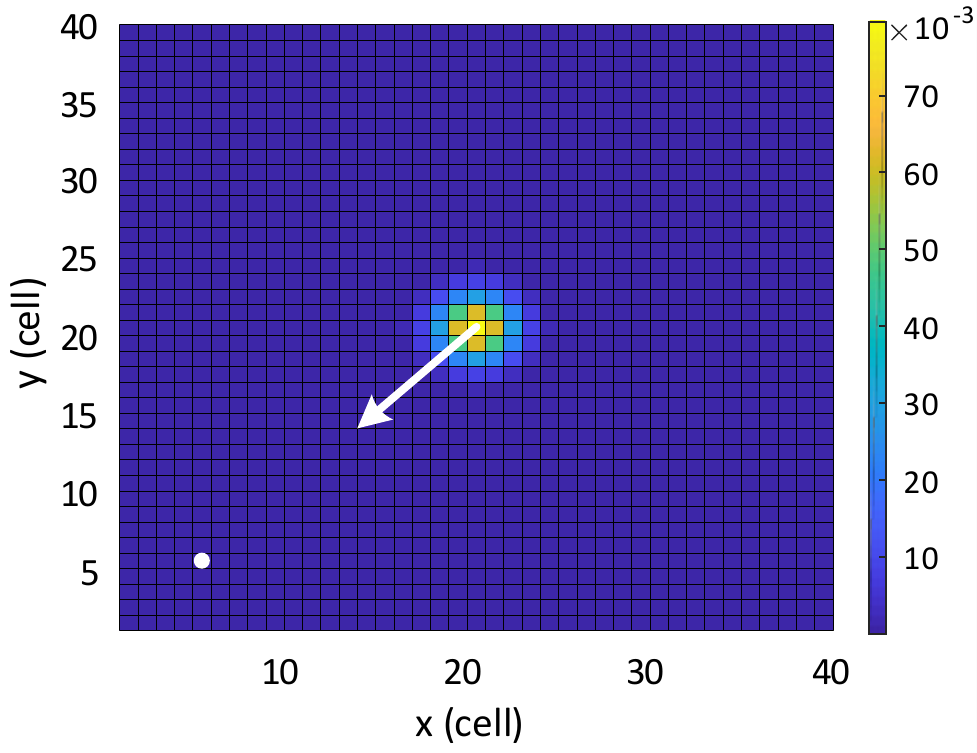}
		\caption{Scenario 4}
		\label{fig:scenario4}
	\end{subfigure}
	\begin{subfigure}{0.5\textwidth}
		\centering
		\includegraphics[width = \textwidth]{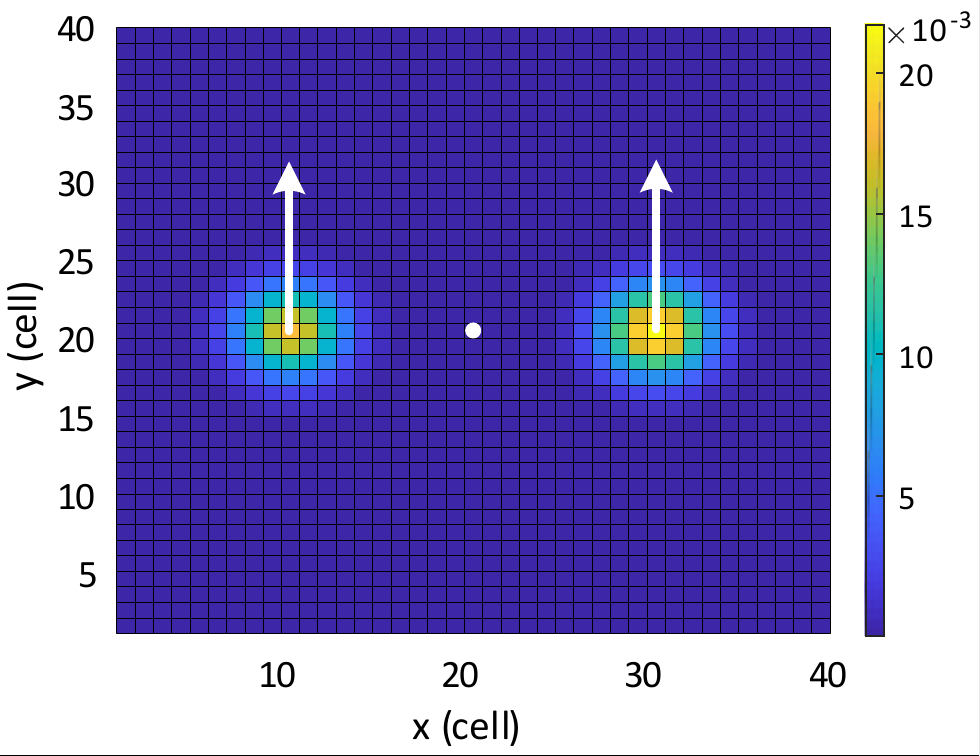}
		\caption{Scenario 5}
		\label{fig:scenario5}
	\end{subfigure}
	\begin{subfigure}{0.5\textwidth}
		\centering
		\includegraphics[width = \textwidth]{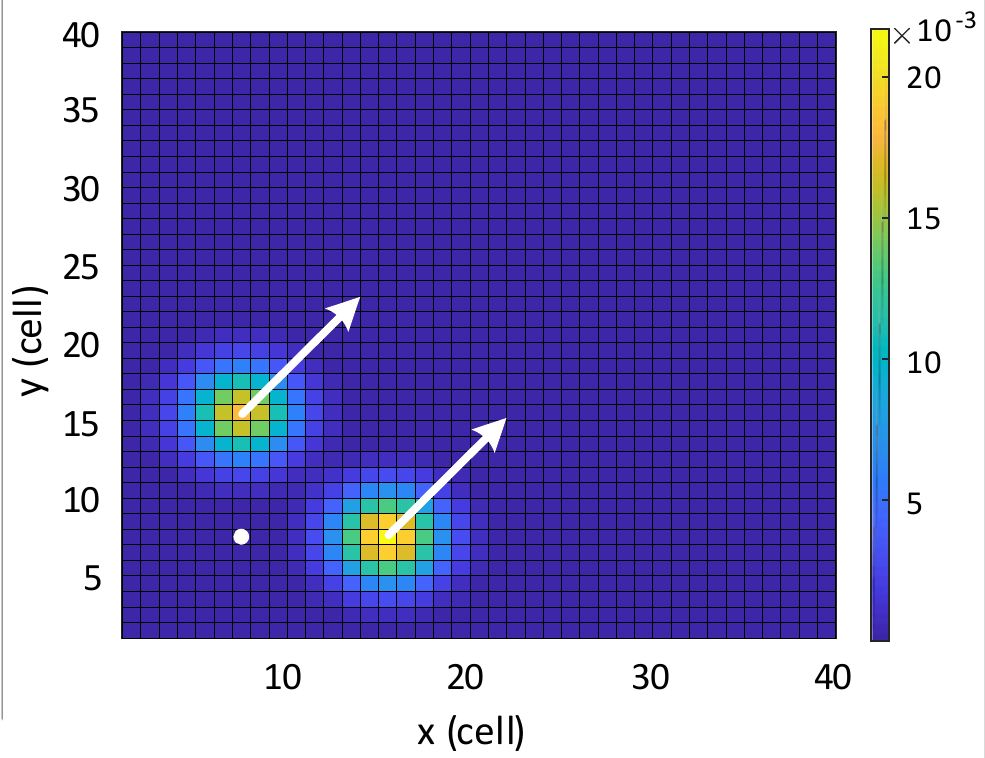}
		\caption{Scenario 6}
		\label{fig:scenario6}
	\end{subfigure}		
	
	\centering
	\caption{Scenarios used for evaluating the searching algorithms}
	\label{fig:scenarios}
\end{figure*}

For the sake of coverage, six different search scenarios are used to analyze the performance of MPSO for optimal search (some of them are adopted from \cite{Sara2018}). The scenarios are defined to have the same map size ($S_r= S_c= 40$), but differ in the initial locations of UAV, target motion model $P(x_t|x_{t-1})$ and initial belief map $b(x_0)$. As shown in Fig.\ref{fig:scenarios}, the probability map is color-coded with the target dynamics presented by a white arrow and the initial location of UAV described by a white circle. The scenarios represent different searching situations as follows:

\textbf{Scenario 1} has two high probability regions located next to each other. They are slightly different in location and value, which may cause difficulty in finding a better region to search for the target.

\textbf{Scenario 2} includes two separated high probability regions located opposite to each other over the UAV location. The algorithm has to quickly identify the higher probability region to search and track as the target is moving south-west.

\textbf{Scenario 3} has one small dense region moving rapidly toward the south-east. It thus tests the algorithm in its exploration and adaptation capability. 

\textbf{Scenario 4} is similar to Scenario 3 except that the target is moving toward the UAV's start location. It further evaluates the adaptability of the searching algorithm. 

\textbf{Scenario 5} consists of two probability regions located oppositely via the start location in which the right region is slightly higher in probability. As the target is moving north, the algorithm needs to identify the correct target region.

\textbf{Scenario 6} is similar to Scenario 5, but the start location is below the potential regions and the target is moving North-East. It thus evaluates the capability of searching in a diagonal direction.

In our evaluations, MPSO is implemented with the parameters $w = 1$ at the damping rate of 0.98, $\varphi_1 = 2.5$ and $\varphi_2 = 2.5$. The swarm size is chosen to be 1000 particles. The number of iterations is 100 and the size of the search path is 20 nodes. Due to the stochastic nature of PSO, the algorithm is executed 10 times to find the average and standard deviation values for each scenario. 

\subsection{Search path}
\begin{figure*}
	
	\begin{subfigure}{0.5\textwidth}
		\centering
		\includegraphics[width=\textwidth]{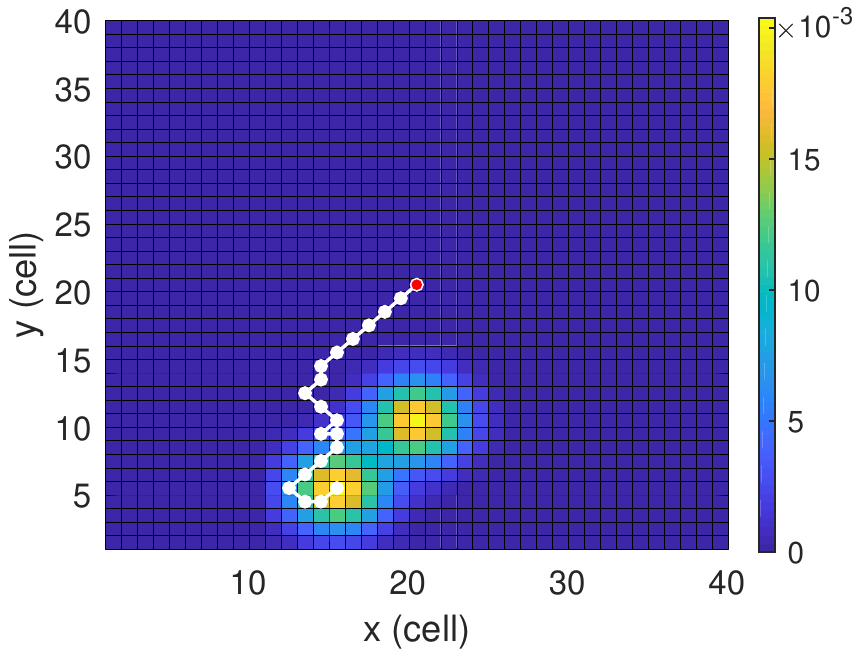}
		\caption{Scenario 1: $P_t = 0.1886$}
		\label{fig:path1}
	\end{subfigure}%
	\begin{subfigure}{0.5\textwidth}
		\centering
		\includegraphics[width=\textwidth]{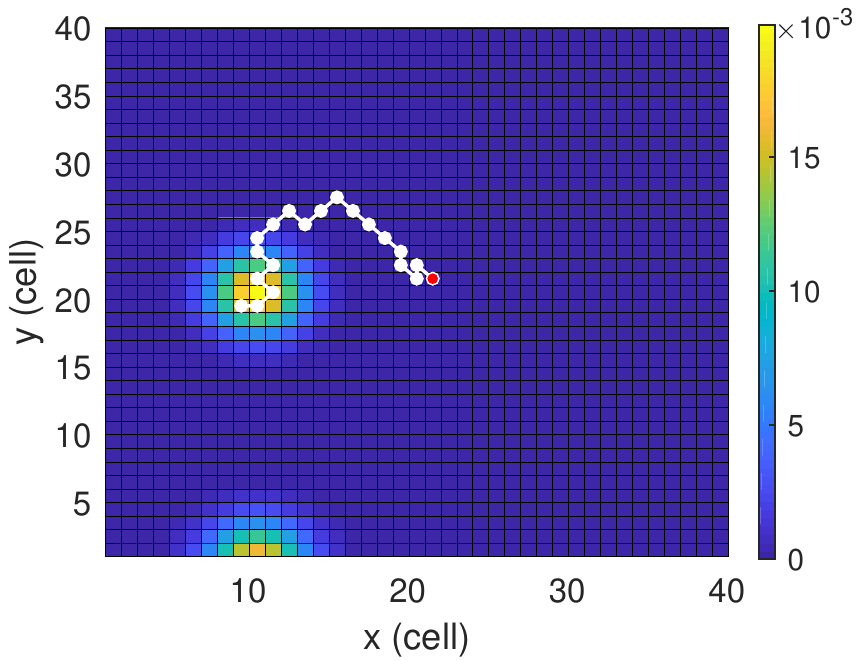}
		\caption{Scenario 2: $P_t = 0.2496$}
		\label{fig:path2}
	\end{subfigure}
	\begin{subfigure}{0.5\textwidth}
		\centering
		\includegraphics[width=\textwidth]{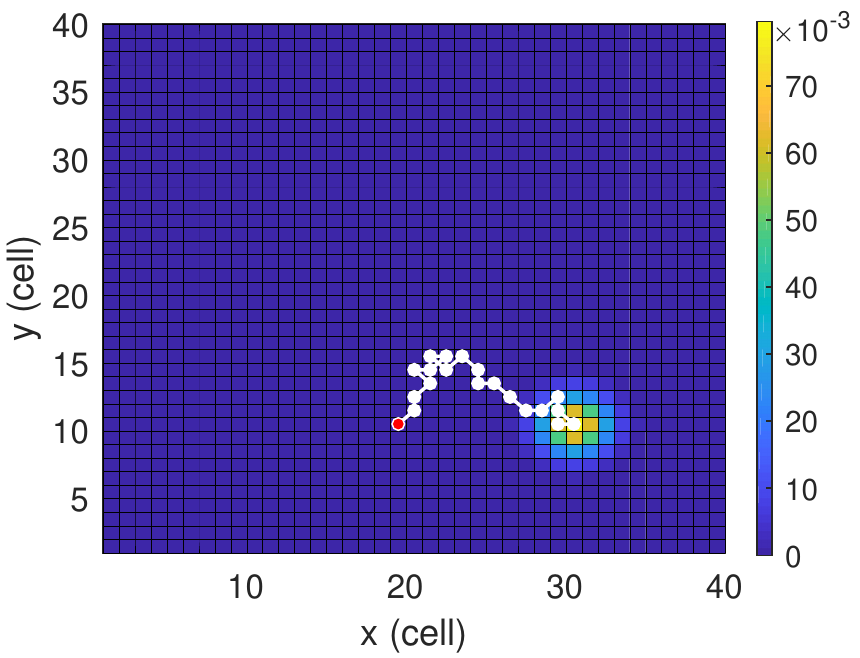}
		\caption{Scenario 3: $P_t = 0.64907$}
		\label{fig:path3}
	\end{subfigure}%
	\begin{subfigure}{0.5\textwidth}
		\centering
		\includegraphics[width=\textwidth]{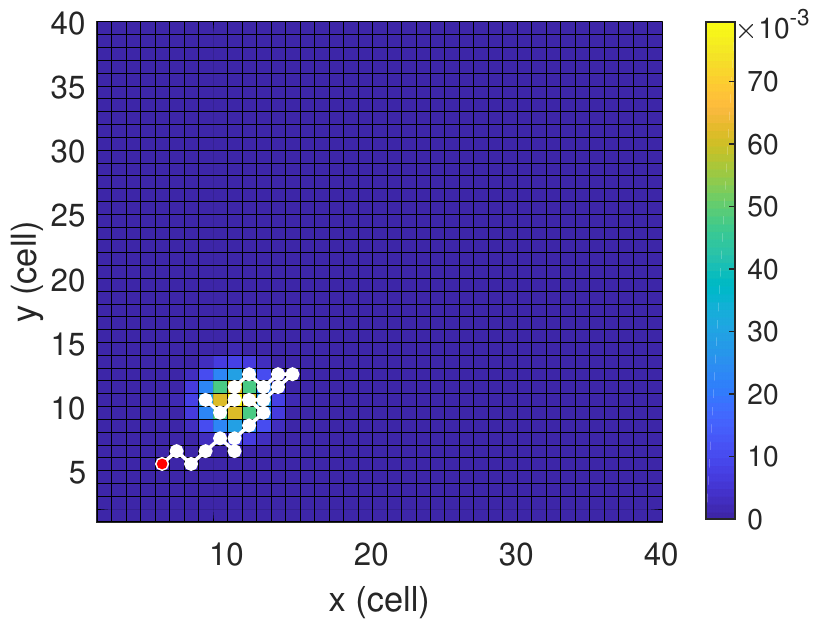}
		\caption{Scenario 4: $P_t = 0.5111$}
		\label{fig:path4}
	\end{subfigure}
	\begin{subfigure}{0.5\textwidth}
		\centering
		\includegraphics[width=\textwidth]{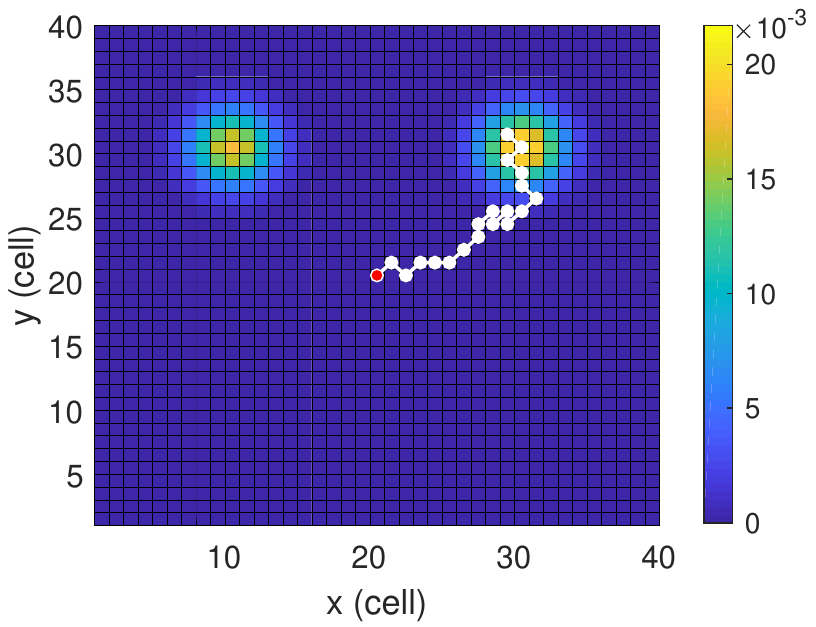}
		\caption{Scenario 5: $P_t = 0.2226$}
		\label{fig:path5}
	\end{subfigure}
	\begin{subfigure}{0.5\textwidth}
		\centering
		\includegraphics[width=\textwidth]{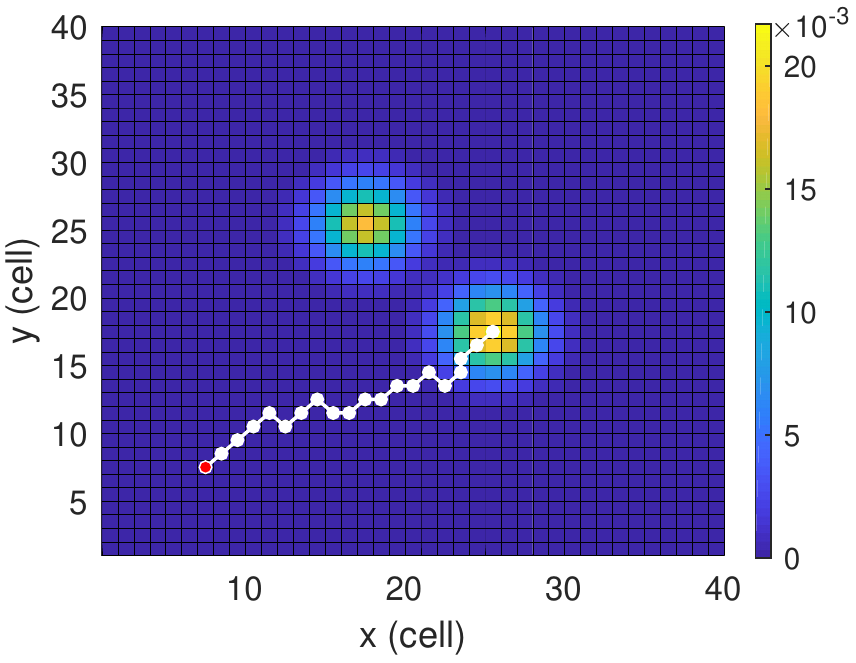}
		\caption{Scenario 6: $P_t = 0.1907$}
		\label{fig:path6}
	\end{subfigure}		
	
	\centering
	\caption{Search paths for each scenario generated by MPSO}
	\label{fig:path}
\end{figure*}

Figure \ref{fig:path} shows the search paths of MPSO for each scenario together with the cumulative probability values. In all scenarios, MPSO is able to find the highest probability regions and generates relevant paths for the UAV to fly. For scenarios with only one high probability region such as Scenario 3 and 4, the cumulative probabilities are high because the chance of finding the target is not spread to other regions. It is also noted from Fig. \ref{fig:path} that the probability map only reflects the target belief at the last step whereas the search path represents the tracking of high probability regions over time. By comparing them with those in Fig. \ref{fig:scenarios}, we can see that the search paths adapt to the target dynamics to maximize the detection probability. 

\begin{table*}
	\centering
	\caption{Comparison between PSO algorithms on fitness representing the accumulated detection probability}
	\label{tab:fitness1}
	\begin{tabular}{cllll}
		\hline
		\rule{0pt}{3ex}
		Scenario & \multicolumn{1}{c}{MPSO} & \multicolumn{1}{c}{PSO} & \multicolumn{1}{c}{QPSO} & \multicolumn{1}{c}{APSO} \\
		\hline
		1        & \textbf{0.1876$\pm$0.0011}   & 0.1476
		$\pm$0.0043    & 0.1198$\pm$0.0037            & 0.1869$\pm$0.0025            \\
		2        & \textbf{0.247$\pm$0.0055}    & 0.2019$\pm$0.0163           & 0.2014$\pm$0.0046            & 0.2393$\pm$0.0113            \\
		3        & 0.6554$\pm$0.014    & 0.5403$\pm$0.0218           & 0.5468$\pm$0.014             & \textbf{0.6649$\pm$0.0287}            \\
		4        & \textbf{0.5018$\pm$0.0095}   & 0.4082$\pm$0.0092           & 0.4259$\pm$0.0164            & 0.4969$\pm$0.0109            \\
		5        & \textbf{0.2213$\pm$0.0025}            & 0.1785$\pm$0.0067           & 0.1819$\pm$0.0008            & 0.2199$\pm$0.004    \\
		6        & \textbf{0.1881$\pm$0.0112}   & 0.097$\pm$0.0239            & 0.0943$\pm$0.0168            & 0.1735$\pm$0.0187           \\
		\bottomrule
	\end{tabular}
\end{table*}

\subsection{Comparison with other PSO algorithms}

We have judged the merit of MPSO over other PSO algorithms including a classical PSO, denoted here as PSO for the comparison purpose, quantum-behaved PSO (QPSO) and angle-encoded PSO (APSO).

\textbf{PSO} is introduced in \cite{Kennedy2001} in which the particles encode a search path as a set of nodes. They then evolve according to (\ref{eq:velocity}) and (\ref{eq:position}) to find the optimal solution.

\textbf{APSO} operates in a similar way as PSO. It, however, encodes the position of particles as a set of phase angles so that each angle represents the direction in which the path would emerge \cite{Fu2012}.

\textbf{QPSO}, on the other hand, assumes particles to have quantum behavior in a bound state. The particles are attracted by a quantum potential well centered on its local attractor and thus have a new stochastic update equation for their positions \cite{Sun2012}. In QPSO, the position of particles also encodes a search path that includes a set of nodes.

\begin{figure*}
	
	\begin{subfigure}{0.5\textwidth}
		\centering
		\includegraphics[width=\textwidth]{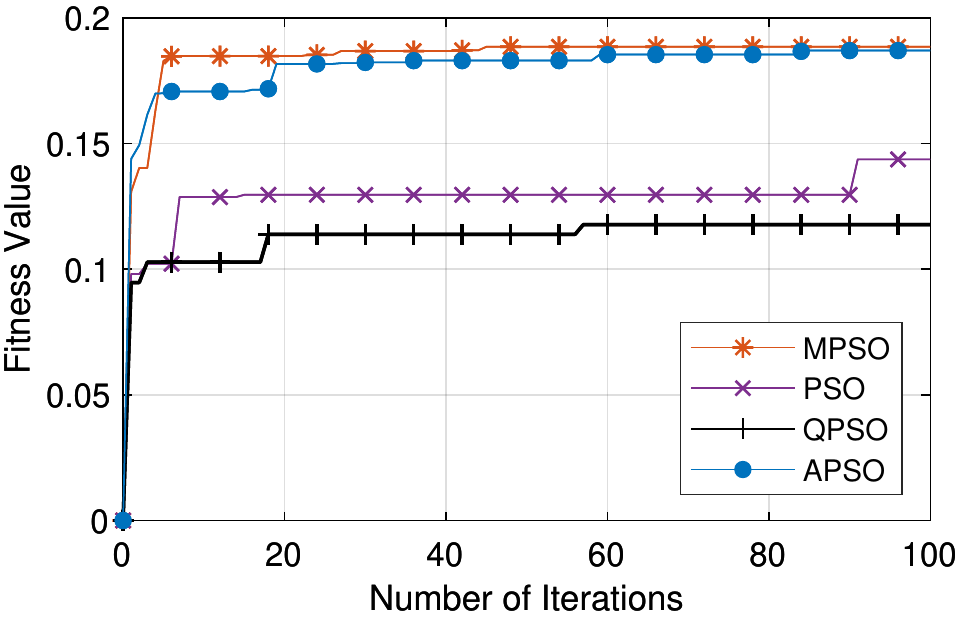}
		\caption{Scenario 1}
		\label{fig:PSO_S1}
	\end{subfigure}%
	\begin{subfigure}{0.5\textwidth}
		\centering
		\includegraphics[width=\textwidth]{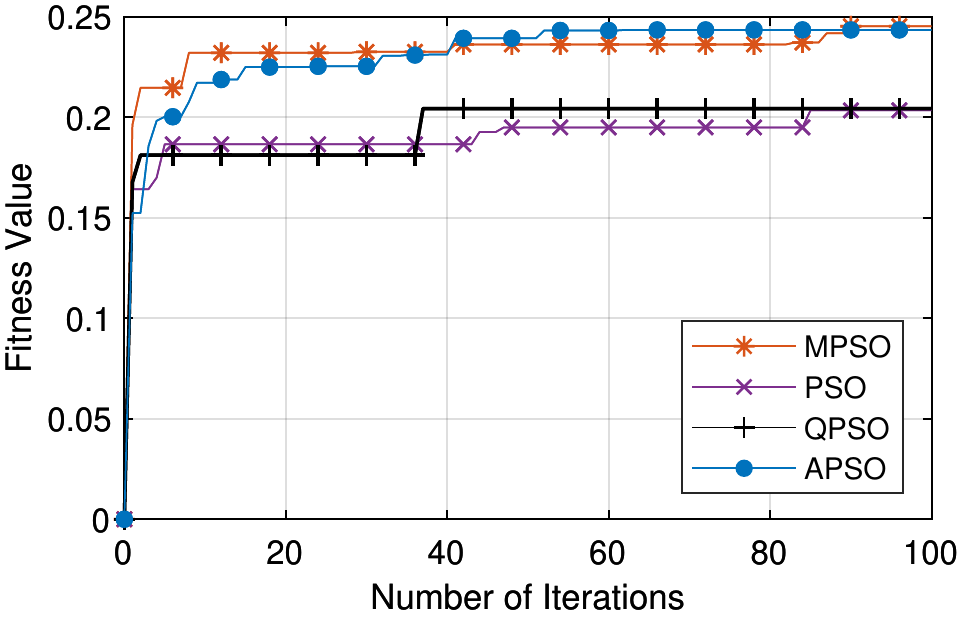}
		\caption{Scenario 2}
		\label{fig:PSO_S2}
	\end{subfigure}
	\begin{subfigure}{0.5\textwidth}
		\centering
		\includegraphics[width=\textwidth]{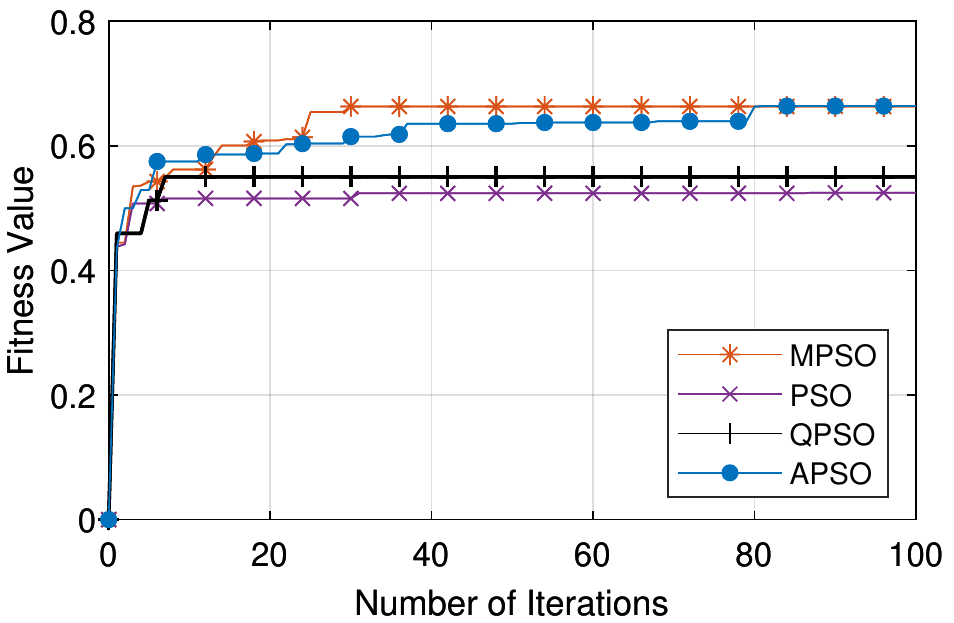}
		\caption{Scenario 3}
		\label{fig:PSO_S3}
	\end{subfigure}%
	\begin{subfigure}{0.5\textwidth}
		\centering
		\includegraphics[width=\textwidth]{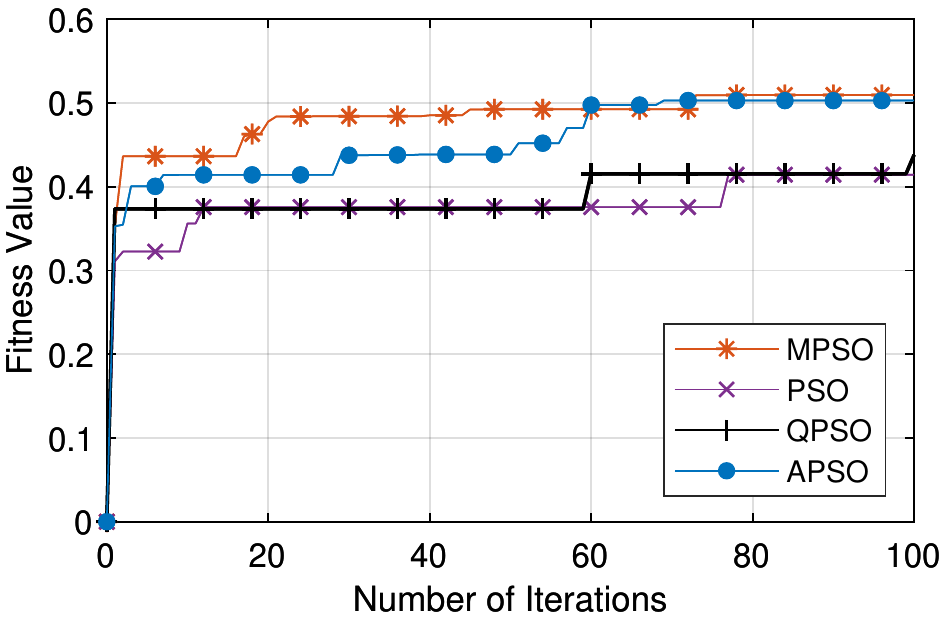}
		\caption{Scenario 4}
		\label{fig:PSO_S4}
	\end{subfigure}
	\begin{subfigure}{0.5\textwidth}
		\centering
		\includegraphics[width=\textwidth]{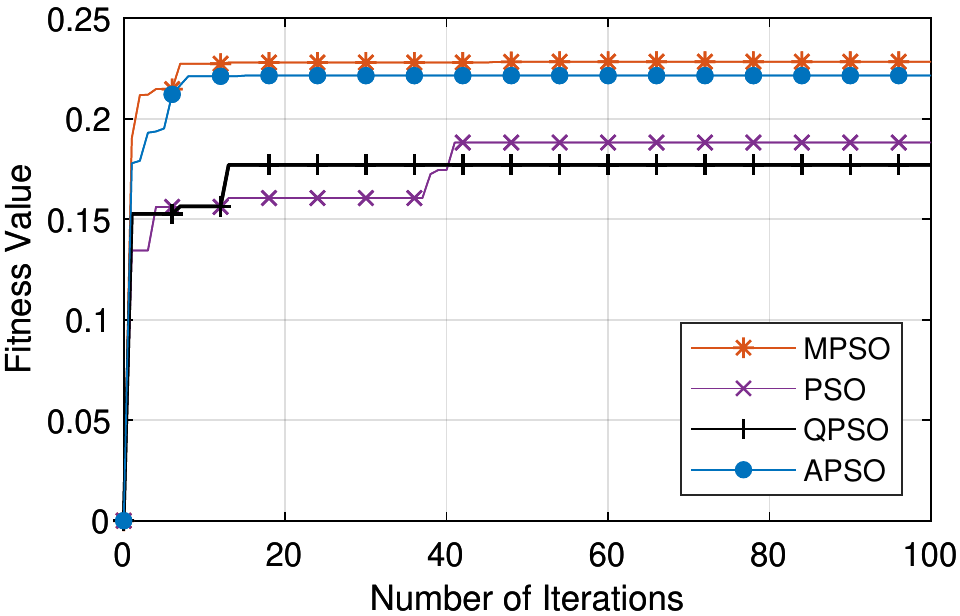}
		\caption{Scenario 5}
		\label{fig:PSO_S5}
	\end{subfigure}
	\begin{subfigure}{0.5\textwidth}
		\centering
		\includegraphics[width=\textwidth]{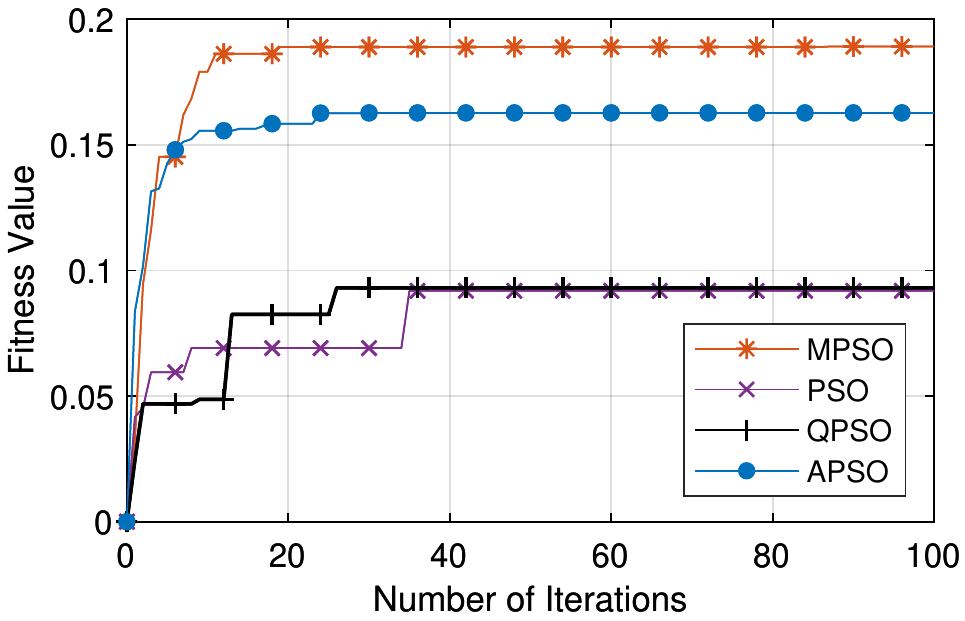}
		\caption{Scenario 6}
		\label{fig:PSO_S6}
	\end{subfigure}		
	
	\centering
	\caption{Convergence curves of the four PSO algorithms on the six benchmark scenarios}
	\label{fig:PSO_Convergence}
\end{figure*}

Table \ref{tab:fitness1} shows the average and standard deviation values of the fitness representing the accumulated detection probability obtained by all algorithms after 10 runs. It can be seen that MPSO introduces the best performance in 5 scenarios. APSO is slightly better than MPSO in Scenario 3, but its convergence is not stable reflected via a larger standard deviation value. These results can be further verified via the convergence curves shown in Fig. \ref{fig:PSO_Convergence}. They show that PSO and QPSO present poor performance as the use of nodes to encode search paths does not maintain particle momentum resulting in local maxima. 

APSO, on the other hand, introduces a comparable performance with MPSO. Unlike PSO and QPSO, the use of angles in APSO allows particles to search in orientation space and thus maintains the swarm properties. Interestingly, APSO can be considered as a special case of MPSO when the motion magnitude is constrained to 1. While this constraint limits the flexibility of the swarm, it may improve the exploration capacity in certain scenarios to yield a good result such as in Scenario 3.  

\subsection{Comparison with metaheuristic optimization algorithms}
To further evaluate the performance of MPSO, we have compared it with state-of-the-art metaheuristic optimization algorithms including the artificial bee colony (ABC), ant colony optimization (ACO), genetic algorithm (GA), differential evolution (DE), and tree-seed algorithm (TSA).

\begin{table*}
	\centering
	\caption{Comparison between MPSO and other metaheuristic algorithms on fitness}
	\label{tab:fitness2}
	\begin{tabular}{cllllll}
		\hline
		\rule{0pt}{3ex}
		Scenario & \multicolumn{1}{c}{MPSO} & \multicolumn{1}{c}{ABC} & \multicolumn{1}{c}{GA} & \multicolumn{1}{c}{ACO} & \multicolumn{1}{c}{DE} & \multicolumn{1}{c}{TSA} \\
		\hline
		1        & \textbf{0.1876$\pm$0.0011}   & 0.1691$\pm$0.0076           & 0.1283$\pm$0.0001          & 0.1836$\pm$0.0013 & 0.1818$\pm$0.0015 & 0.1873$\pm$0.0006  \\
		2        & \textbf{0.247$\pm$0.0055}    & 0.2099$\pm$0.0041           & 0.2151$\pm$0.0018          & 0.2145$\pm$0.0049 & 0.22$\pm$0.0045 & 0.2362$\pm$0.0085    \\
		3        & \textbf{0.6554$\pm$0.014}    & 0.5872$\pm$0.0152           & 0.5995$\pm$0.003           & 0.6053$\pm$0.02 & 0.5985$\pm$0.0166 & 0.6236$\pm$0.0135             \\
		4        & \textbf{0.5018$\pm$0.0095}   & 0.4225$\pm$0.0017           & 0.3497$\pm$0.0311          & 0.4866$\pm$0.0139 & 0.4243$\pm$0.0252 & 0.4626$\pm$0.0239           \\
		5        & \textbf{0.2213$\pm$0.0025}   & 0.2093$\pm$0.0071           & 0.1733$\pm$0.0001          & 0.2208$\pm$0.0024 & 0.2128$\pm$0.006 & 0.2209$\pm$0.0005           \\
		6        & 0.1881$\pm$0.0112   & 0.181$\pm$0.0019            & 0.1255$\pm$0.0001          & 0.15$\pm$0.0119 & 0.1829$\pm$0.0139 & \textbf{0.1889$\pm$0.0018}   \\
		\bottomrule		 
	\end{tabular}
\end{table*}

\textbf{ABC} searches for optimal solutions based on the cooperative behavior of three types of bees: employed bees, onlooker bees and scout bees \cite{Karaboga2008}. Our implementation represents each solution as a search path that consists of a set of motion segments similar to MPSO.

\textbf{ACO} solves optimization problems based on heuristic information and a pheromone model of artificial ants, each maintains a feasible solution \cite{Dorigo2003}. Our implementation of ACO is based on \cite{Sara2018} in which the ``ACO-Node+H" approach is used together with the max-min ACO.   

\textbf{GA} is a popular metaheuristic optimization that modifies a population of individual solutions similar to the process of natural selection \cite{Goldberg1989}. Our implementation of GA is based on the ``EA-dir" approach in \cite{Lin2009} where a path is encoded as a string of directions subjected to two mutation techniques including ``flip" and ``pull".

\textbf{DE} is an optimization method that finds the optimal solution by improving its candidates via simple mathematical formulas from a population of individual solutions \cite{storn1997}. In implementing DE for optimal search, we represent each solution as a set of motions similar to the representation used in MPSO.

\textbf{TSA} solves the optimization problem by simultaneously exploring and exploiting the search space based on the spread of seeds from a tree population. The level and balance between the exploration and exploitation are controlled by predefined parameters including the search tendency ($ST$) and the number of seeds ($NS$). Those parameters are chosen as in the original study \cite{KIRAN2015} in our implementation, i.e., $ST = 0.1$ and $ NS \in [0.1,0.25]$.

Table \ref{tab:fitness2} presents the fitness values corresponding to the optimal solutions of MPSO and metaheuristic algorithms over six scenarios after 10 runs. The values include the average and standard deviation representing the cumulative detection probability. It can be seen that MPSO outperforms other metaheuristic algorithms in scenarios 1 to 5 with the highest fitness values and small standard deviation. TSA is the second best with satisfactory results in most scenarios, whereas the remaining algorithms are only good in one or two scenarios. 

\begin{figure*}
	
	\begin{subfigure}{0.5\textwidth}
		\centering
		\includegraphics[width=\textwidth]{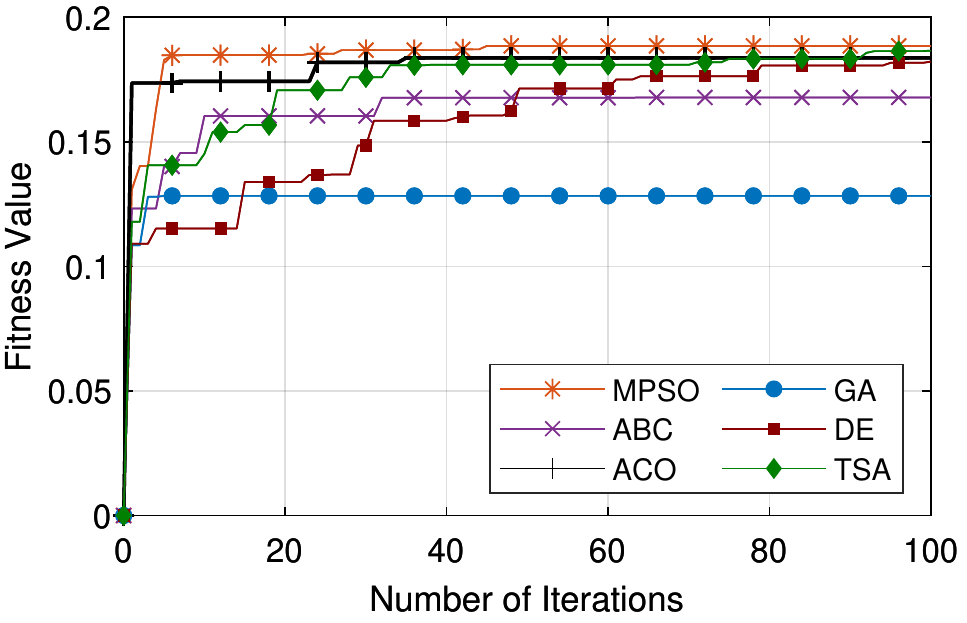}
		\caption{Scenario 1}
		\label{fig:PSOACO_S1}
	\end{subfigure}%
	\begin{subfigure}{0.5\textwidth}
		\centering
		\includegraphics[width=\textwidth]{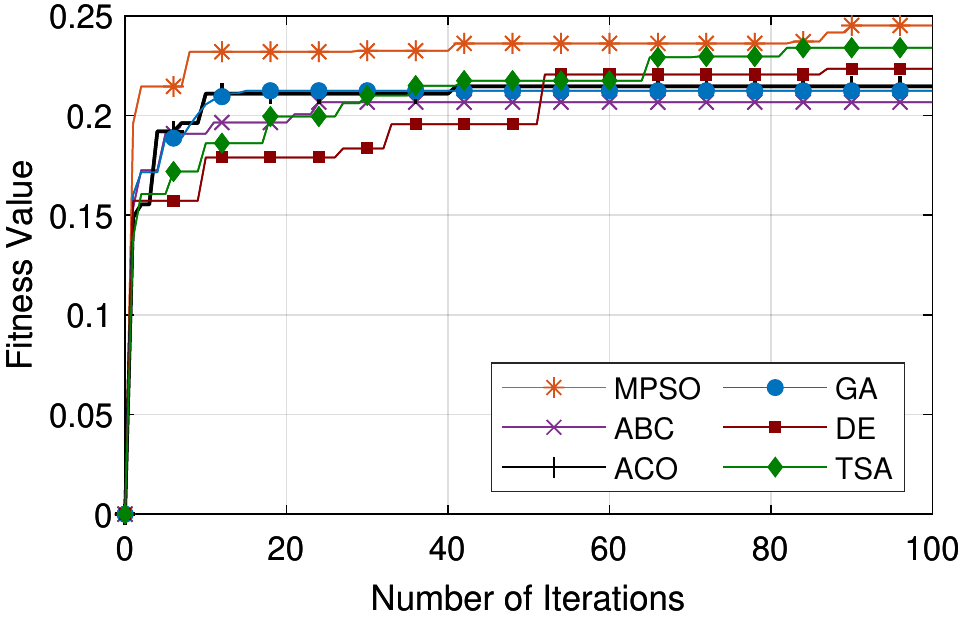}
		\caption{Scenario 2}
		\label{fig:PSOACO_S2}
	\end{subfigure}
	\begin{subfigure}{0.5\textwidth}
		\centering
		\includegraphics[width=\textwidth]{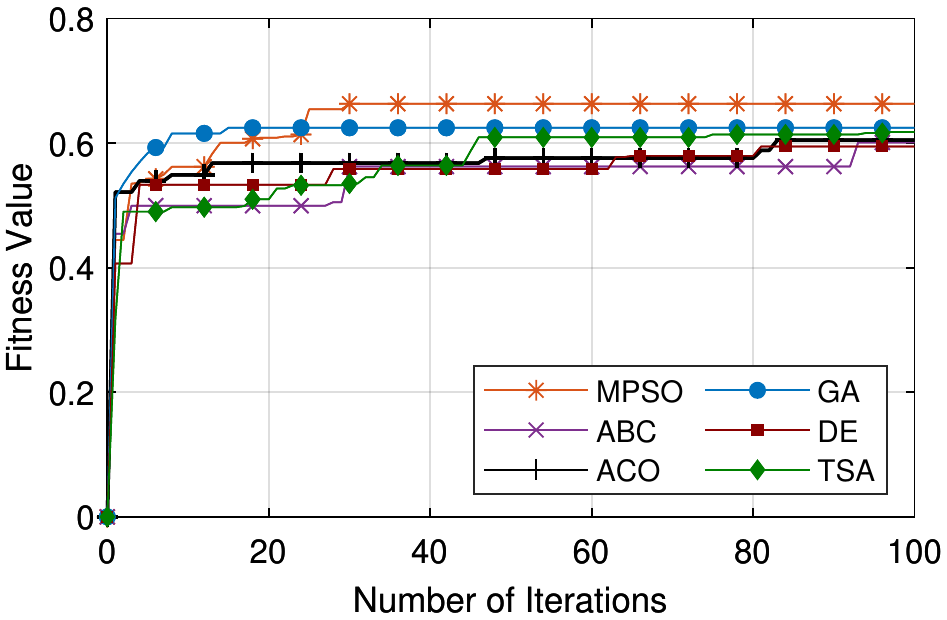}
		\caption{Scenario 3}
		\label{fig:PSOACO_S3}
	\end{subfigure}%
	\begin{subfigure}{0.5\textwidth}
		\centering
		\includegraphics[width=\textwidth]{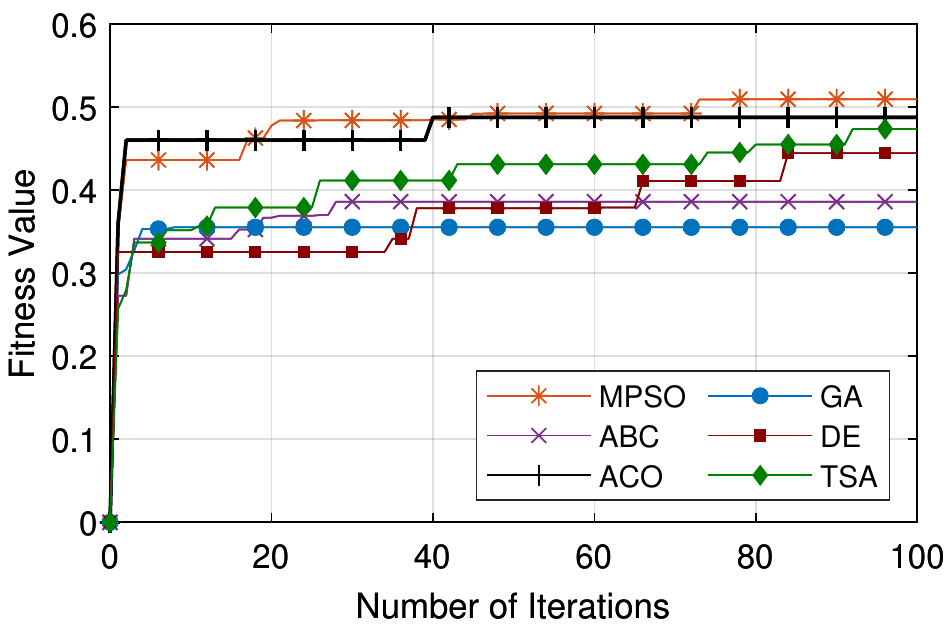}
		\caption{Scenario 4}
		\label{fig:PSOACO_S4}
	\end{subfigure}
	\begin{subfigure}{0.5\textwidth}
		\centering
		\includegraphics[width=\textwidth]{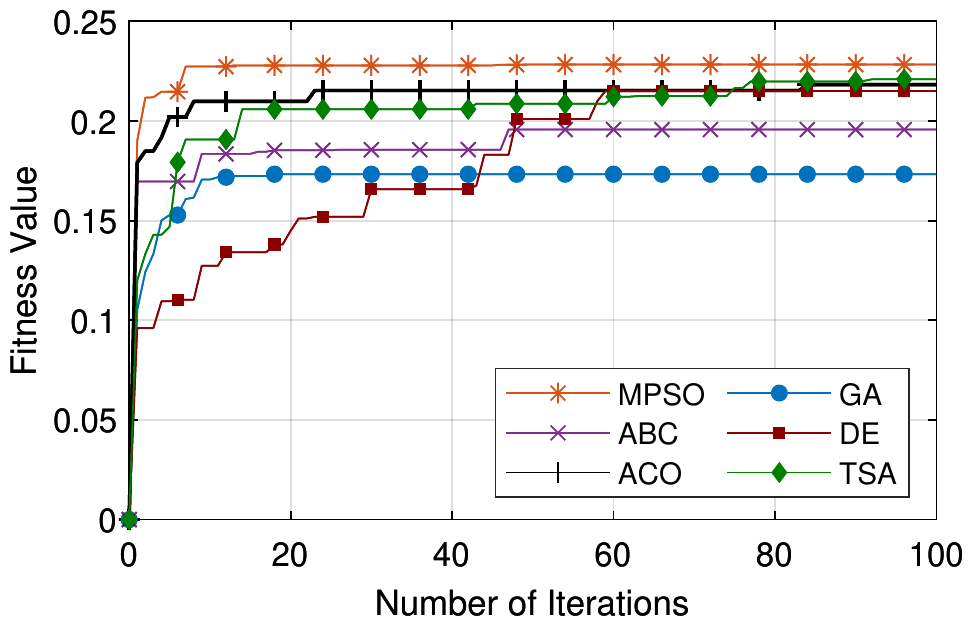}
		\caption{Scenario 5}
		\label{fig:PSOACO_S5}
	\end{subfigure}
	\begin{subfigure}{0.5\textwidth}
		\centering
		\includegraphics[width=\textwidth]{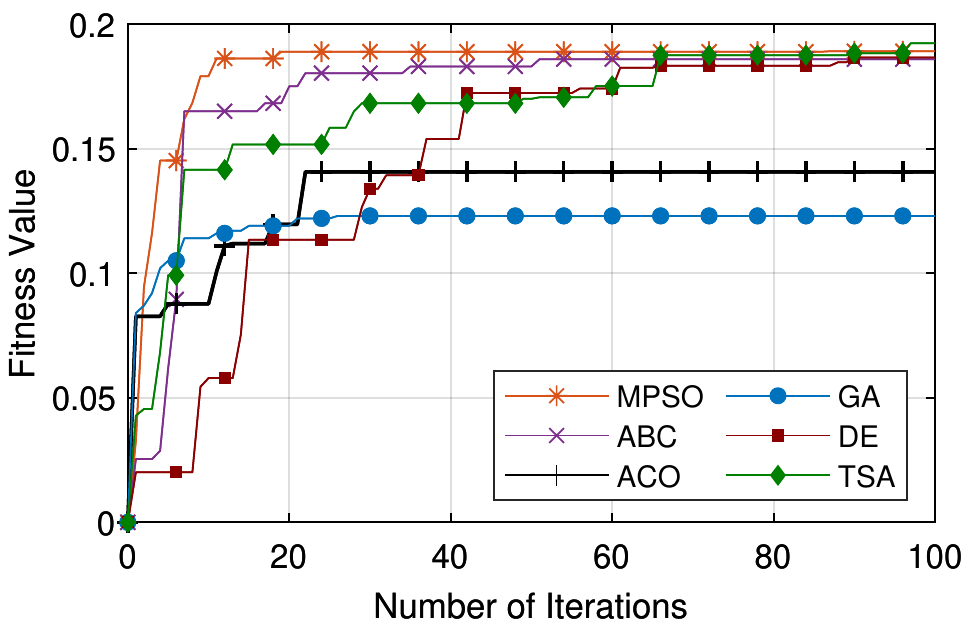}
		\caption{Scenario 6}
		\label{fig:PSOACO_S6}
	\end{subfigure}		
	
	\centering
	\caption{Convergence curves of MPSO and other metaheuristic algorithms on the six benchmark scenarios}
	\label{fig:PSOACO_Convergence}
\end{figure*}

Figure \ref{fig:PSOACO_Convergence} further compares the convergence among the algorithms. While MPSO shows good exploitation capability represented via the high fitness value in most scenarios, its exploration reflected via the convergence speed is rather slow in some scenarios such as Scenario 3 where the high probability region is small and the target is moving away from the UAV. TSA, on the other hand, is good at exploration but rather limited in exploitation so that its final fitness values are slightly less than MPSO. ACO performs well in detecting static and slow-moving targets, but its adaptation to fast-moving targets is limited due to the nature of ACO incrementally exploring via nodes. DE and ABC have stable performance in most scenarios. GA, on the other hand, is often trapped at local minimums as the crossover and mutation operators cause many invalid paths during operation. Besides, the enhanced ``flip" and ``pull" operators which prioritize horizontal and vertical search do not perform well in scenarios requiring diagonal search such as Scenario 6.   

\subsection{Execution time}
Apart from the accuracy, we also evaluate the execution time of all algorithms to roughly estimate their complexity. We executed all algorithms under the same conditions of software and computer hardware. Table \ref{tab:time} shows the average execution time together with the standard deviation after 10 runs on an Intel Core i7-7600U 2.80 GHz processor. It can be seen that MPSO is the fastest in four scenarios, followed by ABC with two scenarios. DE also introduces relatively short execution time due to its simplicity in the search mechanism. TSA, on the other hand, is rather slow due to the extra computation required to evaluate the seeds of each tree. ACO is the slowest because of a large time spent on calculating heuristic information \cite{Sara2018}. Notably, the execution time of APSO is close to MPSO which further explains it as a special case of MPSO. PSO and QPSO both require extra execution time due to the invalid paths generated during operation.     

\subsection{Validation on UAV platform}
To demonstrate the practical use of MPSO, we have applied it to real searching scenarios with details as follows.

\begin{table*}
	\centering
	\caption{Comparison between MPSO and other algorithms on execution time in seconds}
	\label{tab:time}
	\begin{tabular}{clllllllll}
		\hline
		\rule{0pt}{3ex}
		Scenario & \multicolumn{1}{c}{MPSO} & \multicolumn{1}{c}{PSO} & \multicolumn{1}{c}{QPSO} & \multicolumn{1}{c}{APSO} & ABC           & GA   & ACO & DE   & TSA   \\
		\hline
		1        & 43$\pm$2                     & 129$\pm$6                   & 140$\pm$15                   & 50$\pm$8                     & \textbf{34$\pm$1} & 85$\pm$2 & 144$\pm$3 & 37$\pm$3 & 84$\pm$2 \\
		2        & \textbf{26$\pm$4}            & 150$\pm$7                   & 180$\pm$22                   & 34$\pm$4                     & 34$\pm$5          & 95$\pm$3 & 157$\pm$2  & 32$\pm$6 & 57$\pm$6\\
		3        & \textbf{30$\pm$8}           & 142$\pm$4                   & 149$\pm$3                    & 39$\pm$4                     & 31$\pm$4          & 97$\pm$1 & 150$\pm$5  & 34$\pm$2 & 50$\pm$2\\
		4        & \textbf{20$\pm$2}            & 149$\pm$7                   & 149$\pm$1                    & 32$\pm$5                     & 30$\pm$3          & 92$\pm$3 & 133$\pm$3  & 26$\pm$3 & 47$\pm$1\\
		5        & \textbf{29$\pm$7}           & 126$\pm$4                   & 129$\pm$5                    & 46$\pm$5                     & 34$\pm$4          & 92$\pm$3 & 150$\pm$4  & 31$\pm$3 & 60$\pm$5\\
		6        & 48$\pm$7                     & 140$\pm$3                   & 139$\pm$2                    & 61$\pm$1                     & \textbf{39$\pm$3} & 99$\pm$2 & 146$\pm$13 & \textbf{39$\pm$3} & 85$\pm$2\\
		\bottomrule		 
	\end{tabular}
\end{table*}

\begin{figure*}	
	\begin{subfigure}{0.5\textwidth}
		\centering
		\includegraphics[width=0.8\textwidth]{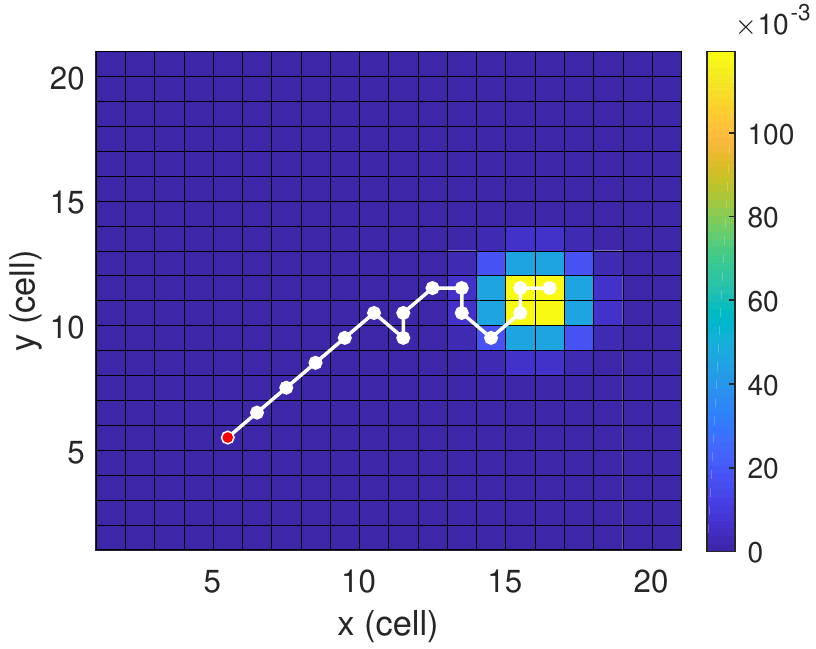}
		\caption{Belief map and search path in experimental scenario 1}
		\label{fig:PlannedPath}
	\end{subfigure}	
	\begin{subfigure}{0.5\textwidth}
		\centering
		\includegraphics[width=0.8\textwidth]{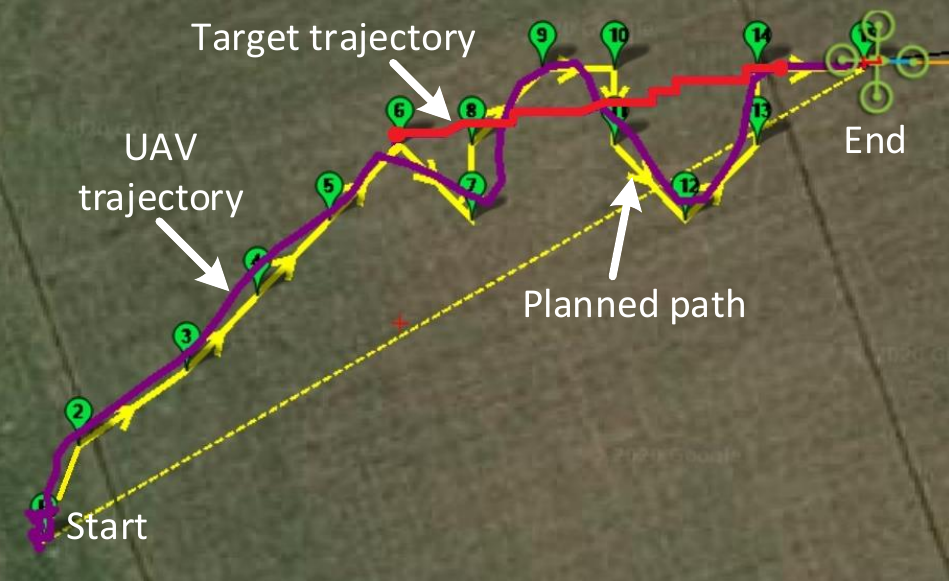}
		\caption{Planned and actual flight paths in experimental scenario 1}
		\label{fig:RealPath}
	\end{subfigure}	
	\begin{subfigure}{0.5\textwidth}
		\centering
		\includegraphics[width=0.8\textwidth]{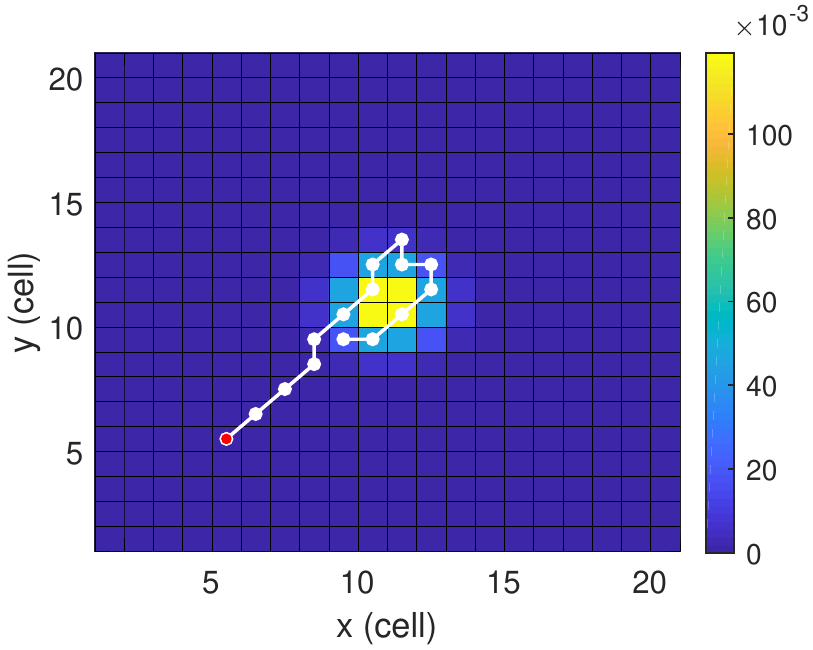}
		\caption{Belief map and search path in experimental scenario 2}
		\label{fig:PlannedPath2}
	\end{subfigure}	
	\begin{subfigure}{0.5\textwidth}
		\centering
		\includegraphics[width=0.8\textwidth]{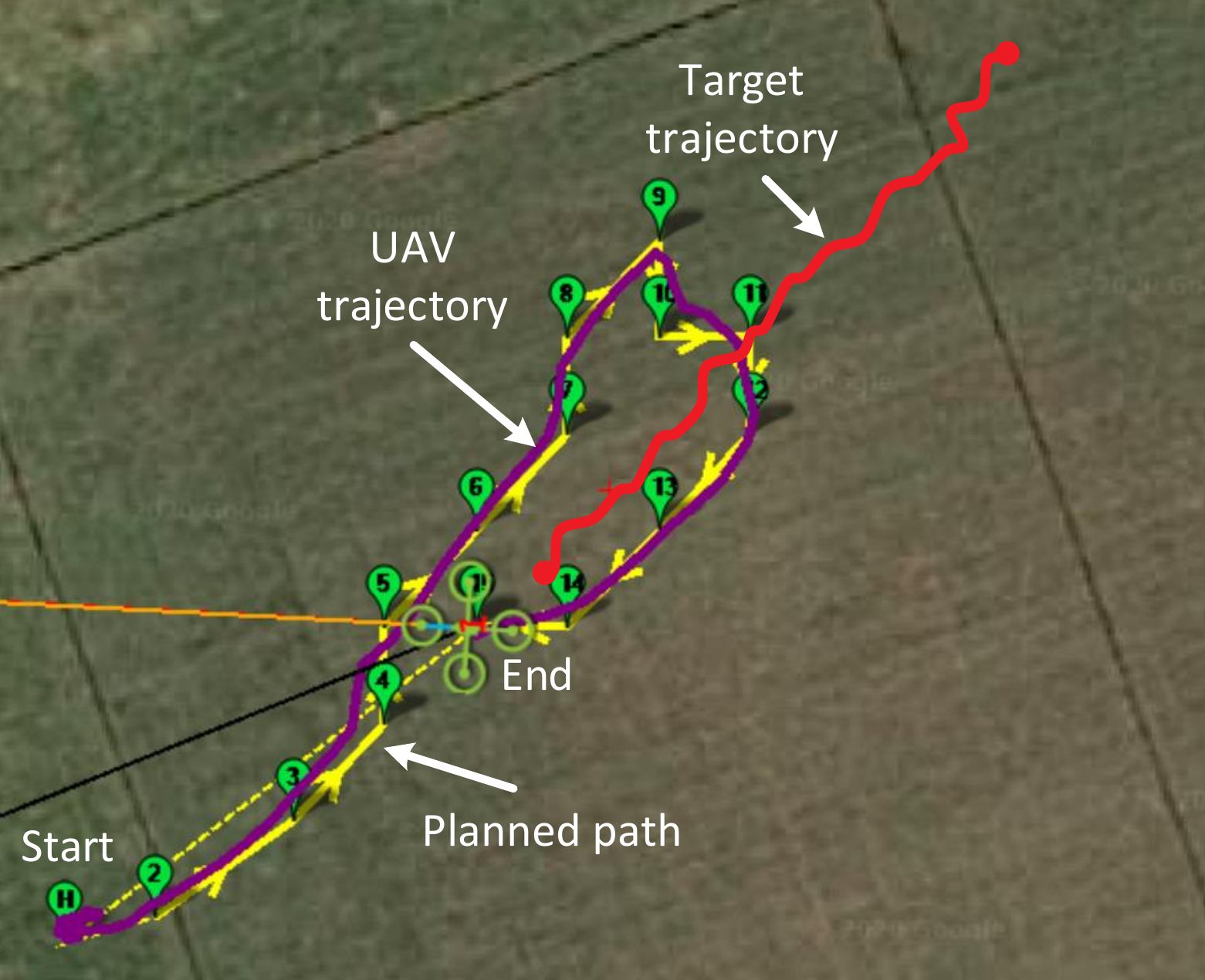}
		\caption{Planned and actual flight paths in experimental scenario 2}
		\label{fig:RealPath2}
	\end{subfigure}	
	
	\centering
	\caption{Experimental detection results }
	\label{fig:FlightPaths}
\end{figure*}

\subsubsection{Experimental setup}
The experiment is carried out in the search area of 60 m $\times$ 60 m located in a park in Sydney. The UAV used is a 3DR Solo drone with a control architecture developed for infrastructure inspection \cite{Hoang2019} that can be controlled via a ground control station (GCS) software named Mission Planner. The detection sensor is a Hero 4 camera attached to the drone via a three-axis gimble responsible for adjusting and stabilizing the camera. An unmanned ground vehicle (UGV) is used as the target. The UGV is equipped with control and communication modules to allow it to track certain trajectories for the sake of experimental verification. 

\begin{figure}
	\centering
	\includegraphics[width=0.35\textwidth]{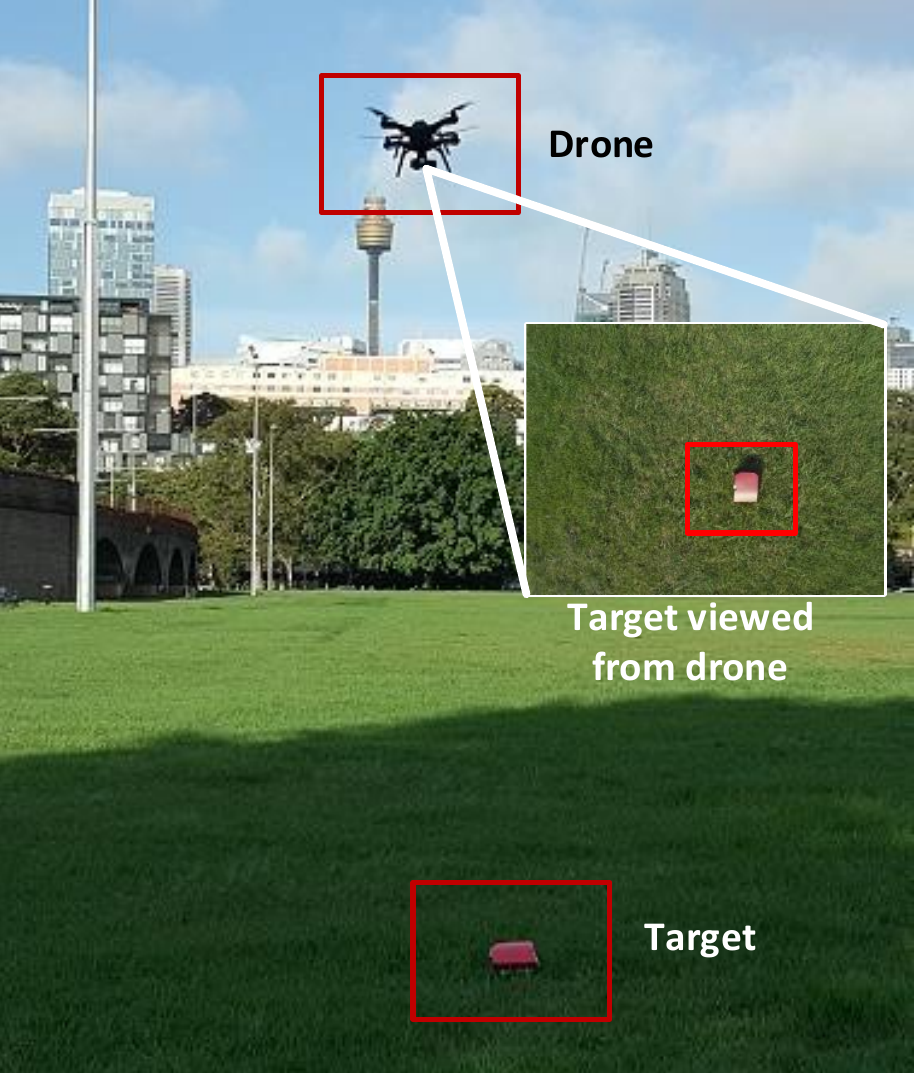}
	\caption{The target within the vision of the camera attached on the drone}
	\label{fig:TargetDetection}
\end{figure}

In experiments, initial locations of UAV and UGV are obtained via the GPS modules equipped on those vehicles and used as the input to generate a belief map. The map is fed to MPSO to generate a search path that includes a list of waypoints. Those waypoints are loaded into Mission Planner to fly the UAV. During the flight, for recording the testing results, positions of the vehicles are tracked via GPS and the video received from the camera is streamed to GCS.

\subsubsection{Experimental results}
Figure \ref{fig:PlannedPath} shows the belief map and path generated by MPSO for the scenario in which the UGV started from the center of the map at the latitude of -33.875992 and the longitude of 151.19145 and moved in East direction. Figure \ref{fig:RealPath} shows the planned and actual flight paths recorded via Mission Planner together with the actual path of UGV. It can be seen that the flight path tracks the planned path with some inevitably small tracking errors caused by GPS positioning. Those errors can be compensated for by extending the field of view of the detection camera via the flight attitude. The UAV thus can trace and approach the target at the location of (-33.87598,151.19153), as shown in Figure \ref{fig:RealPath}. This can be verified in Fig. \ref{fig:TargetDetection} that displays the target within the vision of the camera. 

In another experiment where the UGV moves toward the starting location of the UAV, the planned path adapts to it by turning backward as shown in Fig. \ref{fig:PlannedPath2}. Figure \ref{fig:RealPath2} presents the actual trajectories of the UAV and UGV. It can be seen that the UAV tracks the planned path to approach the target at the location of (-33.875938,151.191515) and then can trace it eventually.  Those results, together with various successful trials, confirm the validity and applicability of our proposed algorithm.

\subsection{Discussion}
Through extensive simulation, thorough comparison and experiments as described above, it can be seen that MPSO presents better performance than other state-of-the-art heuristic algorithms in most search scenarios and is suitable for practical UAV search operations. The rationale for the success of MPSO lies in the motion-encoded mechanism that prevents the algorithm from generating invalid paths during the searching process so that it can avoid the need for re-initialization, and as such, to accelerate the convergence. The motion-encoded mechanism also allows MPSO to search in the motion space instead of the Cartesian space to improve search performance and better adapt to target dynamics. This advantage is clearly reflected in the good search result of MPSO for the challenging Scenario 4 where the target moves in the opposite direction to the search path that requires the UAV to turn around. Nevertheless, like PSO, MPSO may need to increase the swarm size and number of iterations if the search dimension increases \cite{Piccand2008}. In those scenarios, parallel implementation is required to effectively reduce the computation time, and hence, improve the scalability of the proposed algorithm for large-scale systems.

In practical search, the target dynamics may vary depending on the applications so that the deterministic assumption used in this study may go beyond its validity. In those scenarios, a prediction mechanism using optimal estimators such as the Kalman filter \cite{musoff2009} can be employed to provide a prediction of the target trajectory. It is then used to calculate the cumulative probability used in the objective function of MPSO.

\section{Conclusion} \label{conclusion}
We have presented a new algorithm, the motion-encoded particle swarm optimization (MPSO), to solve the problem of optimal search for a moving target using UAVs. The algorithm encodes the search path as a series of motions that are directly applicable to the search problem which constrains the movement of a UAV to its neighbor cells. By changing the search domain from the Cartesian space to motion space, the algorithm is able to adapt to different target dynamics. It also preserves key properties of PSO to enhance the search performance and allows to conduct continuous search in discrete maps. Simulation and experimental results show that the algorithm is effective and practical enough to deploy for search operations. To be effective also for large-scale systems, the proposed algorithm would need parallel computation to further reduce its execution time. Our future work will focus on evaluating MPSO on benchmarking functions and exploring its capability to solve other complex optimization problems.        

\section*{References}
\bibliography{reference}

\end{document}